\documentclass[lettersize,journal]{IEEEtran}
\usepackage{amsmath,amsfonts}
\usepackage{algorithmic}
\usepackage{algorithm}
\usepackage{array}
\usepackage[caption=false,font=normalsize,labelfont=sf,textfont=sf]{subfig}
\usepackage{textcomp}
\usepackage{stfloats}
\usepackage{url}
\usepackage{verbatim}
\usepackage{graphicx}
\usepackage{cite}
\usepackage{booktabs}
\usepackage{tabularx}
\usepackage{amsfonts,amssymb}
\usepackage{multirow}
\usepackage{bbding}
\usepackage[table]{xcolor}
\newcommand{\gr}{\rowcolor[gray]{.95}}

\begin{document}

\title{Dual-former: Hybrid Self-attention Transformer for Efficient Image Restoration }

\author{Sixiang Chen$^{\dag}$,~\IEEEmembership{Student Member,~IEEE}, Tian Ye$^{\dag}$,~\IEEEmembership{Student Member,~IEEE,}\\
Yun Liu$^{\dag}$,~\IEEEmembership{Senior Member,~IEEE}
Erkang Chen$^{*}$,~\IEEEmembership{Senior Member,~IEEE}
        % <-this % stops a space
\thanks{$\dag$ represents the equal contribution. Corresponding authors: Erkang Chen (ekchen@jmu.edu.cn). Sixiang Chen, Tian Ye and Erkang Chen are with School of Ocean Information Engineering, Jimei University. Yun Liu is with College of Artificial Intelligence, Southwest University.}% <-this % stops a space
}

\markboth{Journal of \LaTeX\ Class Files,~Vol.~14, No.~8, August~2021}%
{Shell \MakeLowercase{\textit{et al.}}: A Sample Article Using IEEEtran.cls for IEEE Journals}


\maketitle

\begin{abstract}
Recently, image restoration transformers have achieved comparable performance with previous state-of-the-art CNNs. However, how to efficiently leverage such architectures remains an open problem. In this work, we present Dual-former whose critical insight is to combine the powerful global modeling ability of self-attention modules and the local modeling ability of convolutions in an overall architecture. With convolution-based Local Feature Extraction modules equipped in the encoder and the decoder, we only adopt a novel Hybrid Transformer Block in the latent layer to model the long-distance dependence in spatial dimensions and handle the uneven distribution between channels. Such a design eliminates the substantial computational complexity in previous image restoration transformers and achieves superior performance on multiple image restoration tasks. Experiments demonstrate that Dual-former achieves a 1.91dB gain over the state-of-the-art MAXIM method on the Indoor dataset for single image dehazing while consuming \textit{only} 4.2\% GFLOPs as MAXIM. For single image deraining, it exceeds the SOTA method by 0.1dB PSNR on the average results of five datasets with \textit{only} 21.5\% GFLOPs. Dual-former also substantially surpasses the latest desnowing method on various datasets, with fewer parameters.

\end{abstract}
\begin{IEEEkeywords}
Image Restoration, Local Feature Extraction, Hybrid Self-attention, Adaptive Control Module, Multi-branch Feed-forward Network.
\end{IEEEkeywords}

\section{Introduction}
\IEEEPARstart{R}{ecently}, non-trivial progress has been made in  image restoration via data-driven CNNs methods (image dehazing~\cite{dmt}, desnowing~\cite{hdcwnet}, deraining~\cite{mpr},image denoising~\cite{chen2022simple}, image debluring~\cite{kupyn2019deblurgan} and underwater enhancement~\cite{fu2022underwater}). The local inductive bias and translation invariance of the convolution operation bestow CNNs the advantages of local details modeling. However, CNN-based designs lack global modeling ability. To alleviate this problem, previous methods adopt gradual down-sampling to obtain a larger receptive field and enhance information interaction~\cite{mpr}. Nevertheless, the recovery performance of CNNs is still limited due to the lack of global modeling.
\begin{figure}[!t]
    \centering
    \includegraphics[width=8cm]{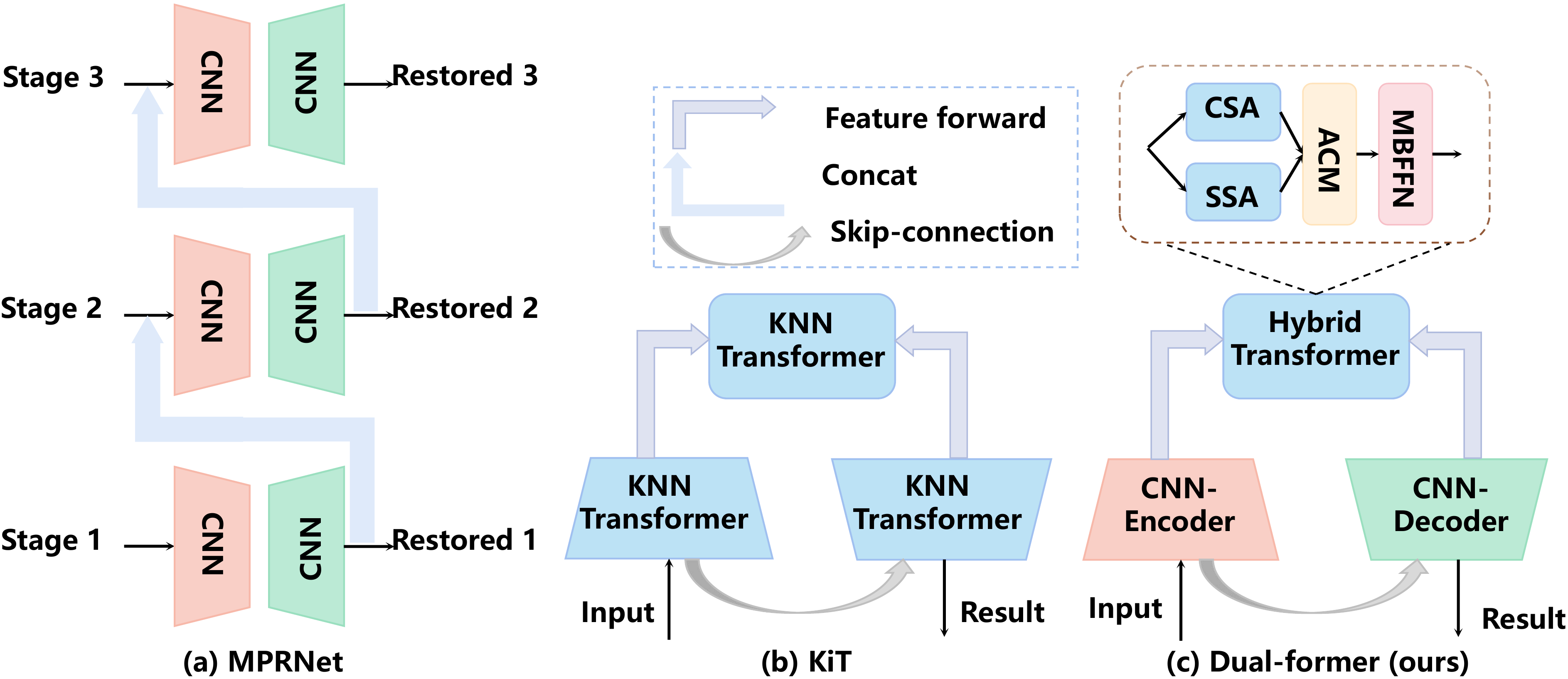}
\end{figure}

\begin{figure}[!t] %通栏
\begin{minipage}[t]{0.33\linewidth} %调节两个子图左右间距
\centering
\includegraphics[width=1\textwidth]{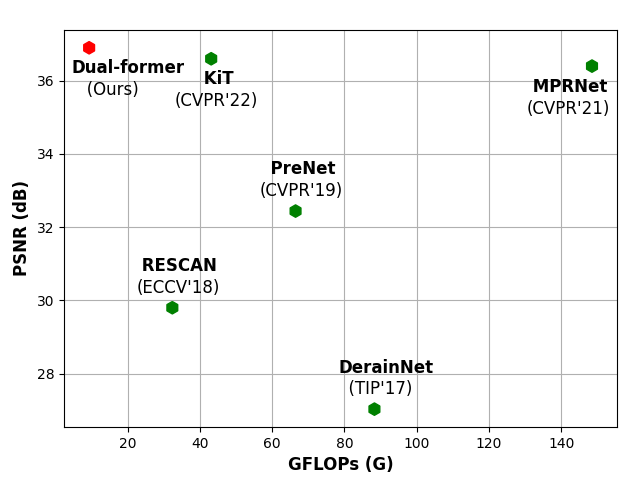} %调节单个子图大小
\label{sots_psnr} %引用标签
\end{minipage}%
\begin{minipage}[t]{0.33\linewidth}
\centering
\includegraphics[width=1\textwidth]{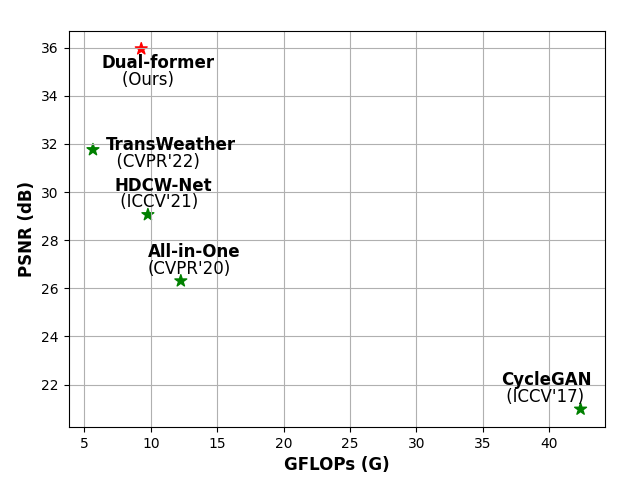}
\label{rain100l_psnr}
\end{minipage}%
\begin{minipage}[t]{0.33\linewidth}
\centering
\includegraphics[width=1\textwidth]{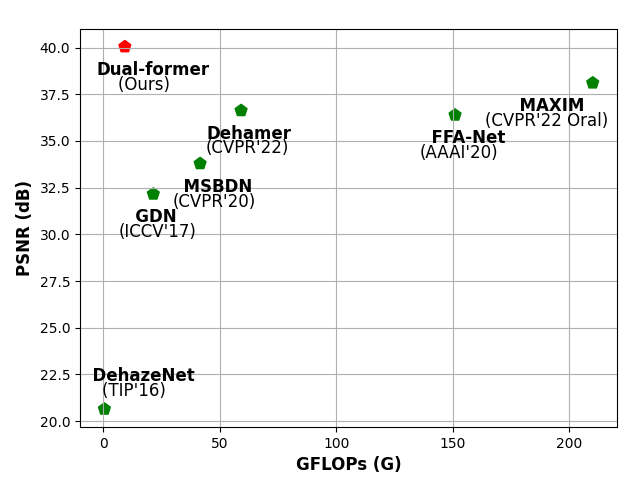}
\label{rain100l_psnr}
\end{minipage}%
\vspace{-0.5mm}
    \caption{\text{Top row:} \textbf{Image Restoration Architecture.} (a) MPRNet~\cite{mpr} designs the multi-stage image restoration backbone, which progressive recovers the degradation image based on CNNs architecture. (b) KiT~\cite{lee2022knn} consists of the k-NN Transformer Block, which utilizes the k-nearest patches obtained by the local-sensitive hashing. (c) Our Dual-former adopts the convolution-based block in encoder-decoder part and embeds the Hybrid Transformer Block into the latent layer to reduce computational complexity. \text{Bottom row:} \textbf{Trade-off between PSNR performance v.s. the number of parameters on Rain100L~\cite{yang2017deep}, CSD~\cite{hdcwnet} and Indoor~\cite{li2018benchmarking} datasets.} Dual-former achieves the best results on various datasets and still costs much less computation complexity than others.}
    \label{fig1}
% \\
% \centering
% \includegraphics[width=45mm]{figure/sots_psnr.png} \label{fig:flops}

\end{figure}

% \begin{figure}[!t]
% \centering
% \begin{minipage}[t]{0.5\linewidth}
% \includegraphics[width=1\textwidth]{figure/sots_psnr.png} \label{fig:parameters}

% \includegraphics[width=1\textwidth]{figure/sots_psnr.png} \label{fig:flops}
% \end{minipage}

% \end{figure}

Inspired by the natural language processing (NLP)~\cite{vaswani2017attention}, vision transformer~\cite{vit} was proposed in the high-level tasks and aroused many wonderful works such as CvT~\cite{wu2021cvt}, Levit~\cite{graham2021levit}, Mobile-former~\cite{mobile-former}, MPViT~\cite{lee2021mpvit}. Vision transformer exploits self-attention to model long-distance dependence, and has shown great potential in low-level vision tasks. 
%Following the above, the third problem to be solved is then raised: (c) \emph{How to save the hardware budget %while utilizing the powerful representation capability of the vision transformer.} 
% Some recent works were trying to reduce the computation. Swim-transformer~\cite{swim} operated self-attention inside the window patch and utilized shifted window partitioning approach to construct connections between neighboring non-overlapping windows. PVT~\cite{PVT} proposed a pyramid vision transformer to build different stages aiming at dropping token numbers. Segformer~\cite{xie2021segformer} used sequence reduction, which adopted linear projection to decrease computational complexity. 

Nevertheless, there is a challenge for ViTs in high-resolution image restoration due to the quadratic relationship between self-attention computation and image size.
To promote the performance and computation trade-off, lots of ViTs-based methods seek careful design to reduce computational burden. Uformer~\cite{wang2022uformer} utilized the window self-attention mechanism combined with the U-shape architecture to limit the amount of computation inside the window. Restormer~\cite{zamir2021restormer} exploited channel-wise self-attention to reduce the computational complexity from square to linear. KiT~\cite{lee2022knn} utilized locality-sensitive hashing, aggregating similar adjacent blocks to improve performance and also restrict computational burden. MAXIM~\cite{tu2022maxim} presented MLP-based modules: the multi-axis block and the cross-gating block, to perform global and local feature interaction with linear complexity.
However, the above methods still have high computational complexities ($\mathbf{89.46G}$ for Uformer, $\mathbf{140.92G}$ for Restormer, $\mathbf{43.08G}$ for KiT and $\mathbf{216.0G}$ for MAXIM in terms of GFLOPs computed in 256$\times$256 resolution). Thus, it is important to \textit{design an effective ViT-based image restoration architecture with lower computational complexity.}

Specifically, previous works such as Restormer~\cite{zamir2021restormer} and KiT~\cite{lee2022knn} used ViT-based modules in the overall framework unsuited for high-resolution features to reduce computational complexity compared with efficient CNNs. While the CNNs-based architecture easily encounters performance bottlenecks. As shown in Fig.\ref{fig1}(a-b), MPRNet~\cite{mpr} based on CNNs still have performance limitation though it has cost substantial computational complexity. Conversely, KiT~\cite{lee2022knn} operates self-attention to model global information and boost the network's performance, while it needs an amount of computational complexity despite reducing computation by locality-sensitive hashing.

According to the study of feature representation~\cite{raghu2021vision}, transformer-based frameworks such as ViT~\cite{vit} primarily pays attention to global information in higher layers and it is hard to perform the local modeling in shallower stages.
In order to combine the merits of CNNs and ViTs, we believe that it's a good paradigm to use convolution-based modules for efficient local modeling while applying self-attention only in the latent layer for global modeling. To this end, we present the Dual-former, which achieves state-of-the-art performance in image restoration tasks with pretty low model complexity (only $\mathbf{9.28G}$ GFLOPs computed
in 256×256 resolution). 
To elevate the efficiency of our framework, (i). we employ the simple yet effective conv-based $\mathbf{L}$ocal $\mathbf{F}$eature $\mathbf{E}$xtraction (LFE) module in the encoder-decoder part, which captures the local information cheaply in high-dimension space. (ii). We only embed the $\mathbf{H}$ybrid $\mathbf{T}$ransformer $\mathbf{B}$lock (HTB) into the latent layer. Compared with previous SOTA methods, this design significantly reduces the computational complexity.

For the Hybrid Transformer Block in the latent layer, we believe long-distance dependence modeling is necessary for image restoration at the spatial level. In addition, the uneven distributions for channel dimensions motivate us that performing sufficient channel global modeling is crucial for image restoration. Thus, we adopt the self-attention mechanism to obtain self-information from both channel and spatial dimensions. Considering the conflict of information from different dimensional perspectives, we propose an $\mathbf{A}$daptive $\mathbf{C}$ontrol $\mathbf{M}$odule (ACM) to fuse the information before implementing a Multi-Branch Feed-forward Network (MBFFN). Parallel to the Hybrid Transformer Block, we leverage our local feature extraction module to augment local information, aiming to enhance the overall representation power in the deepest layer. 
%

% In the part of the experiment, as shown in Fig.\ref{fig1}, we conduct extensive experiments and found that our Dual-former achieves the best results in multiple tasks. For image dehazing, Dual-former outperforms the state-of-the-art methods DMT~\cite{dmt} and MAXIM~\cite{tu2022maxim} 4.32dB and 1.91dB on Haze4K~\cite{dmt} and SOTS-Indoor~\cite{li2018benchmarking} while only use 11.4$\%$ and 4.2$\%$ computation complexity. On the task of removing rain, our model achieves more surprising metrics on average five datasets compared to the state-of-the-art KiT~\cite{lee2022knn} network while keeping the computational load low. We also demonstrate that the Dual-former shows great potential in desnowing task. Although the state-of-the-art method~\cite{valanarasu2022transweather} has slightly less computation complexity, we attract a substantial gain on CSD~\cite{hdcwnet} dataset. $\mathbf{(31.76dB \rightarrow 35.90 dB)}$.

Our contributions can be summarized as follows:
\begin{itemize}
    \item Dual-former combines the merits of CNNs and ViTs via embedding them into different stages to facilitate the trade-off of efficiency and performance.
\end{itemize}
\begin{itemize}
    \item 
    A Local Feature Extraction module (LFE) is proposed to perform local modeling cheaply in high-dimension space via depth-wise convolution and channel attention, which aggregates local information for image restoration.
    % \item In the shallow layer of the network, we design a Local Feature Extraction module (LFE), which can aggregate local information and gradually increase the receptive field through down-sampling in the encoder, providing a better self-attention basis for the Hybrid Transformer Block.
\end{itemize}
\begin{itemize}
    \item We present the Hybrid Transformer Block (HTB), which adaptively controls the information from the hybrid self-attention of the channel and spatial dimensions simultaneously while cooperating with a local convolution block to improve model representation ability in the latent layer.
\end{itemize}
%---------------------------------------------Related Work
\section{Related Works}
\subsection{Image Restoration}
Image restoration is a typical low-level task that aims to restore clean images from degraded images. In recent years, deep learning methods based on convolutional neural networks (CNNs) has widely spread in the field of computer vision, especially in image restoration task, including dehazing~\cite{dmt,ye2021perceiving}, deraining~\cite{mpr,mspfn}, desnowing~\cite{hdcwnet,chen2022snowformer}, denoising~\cite{pang2021recorrupted,huang2022winnet}, debluring~\cite{cho2021rethinking,lee2021iterative} and low light enhancement~\cite{su2020joint,jin2022unsupervised}. UNet~\cite{unet} as a classic encoder-decoder backbone has been utilized in image restoration field, more advanced approaches followed this multi-scale design to process the degraded image version~\cite{abuolaim2020defocus,kupyn2019deblurgan,jin2021dc}. And inspired by other tasks, attention operation is often introduced to image restoration to make up for the lack of interactive information in a certain dimension~\cite{li2018recurrent,dai2019second}, such as channel dimension~\cite{hu2018squeeze} or spatial dimension~\cite{zhao2018psanet}.
With the development of vision transformers, the superiority of global modeling in vision transformers has been migrated to image restoration such as window-based self attention~\cite{liang2021swinir}, self-attention from channel interaction~\cite{zamir2021restormer}, and multi-scale Swin-transformer for image dehazing~\cite{dehazenet}. To learn more image restoration methods, NTIRE challenge reports~\cite{li2022ntire} can be referred.

\subsection{Vision Transformers}
With the efforts of researchers in the visual community, the transformer mechanism has been successfully applied to the field of computer vision~\cite{vit} (ViT), which competes with the CNNs that have dominated for many years. Numerous excellent transformer backbone has been proposed based on it~\cite{PVT,swim,mobile-former}. However, the quadratic computational complexity of the image size is a limitation for ViTs' progress. PVT~\cite{PVT} embedded the transformer block into the pyramid structure, which can gradually reduce the feature size and increase the receptive field to benefit for dense prediction tasks. Swin Transformer~\cite{swim} deliberated drawbacks of ViT and performed self-attention between tokes inside the windows, which kept the linear computation overhead. In addition, it also employed window shift operation to improve communication of window bridging. %MetaFormer~\cite{yu2022metaformer} explored various token mixer methods and presented a pooling operation to replace self-attention or spatial MLP, which did not need a quadratic computational cost of the token number, while the performance shows competitive results.
Many extraordinary architectures like these exemplary methods could be read carefully in ~\cite{liu2022swin,mehta2021mobilevit}.

In low-level vision tasks, more and more transformer-based architectures are introduced. Uformer~\cite{wang2022uformer} was based on Swin Transformer to form a U-shaped architecture for image restoration and shows magnificent performance.
%For single image dehazing, Dehamer ~\cite{Guo_2022_CVPR} designed a novel 3D-position embedding %for single image dehazing, improving the dehazing effect.
KiT~\cite{lee2022knn} grouped k-nearest neighbor patches with locality-sensitive hashing (LSH), which outperformed the state-of-the-art approaches in image deraining. 

Although ViTs have achieved impressive results in low-level vision tasks. However, since the above manners consist of single-type transformer-based blocks in general, its enormous computational complexity makes it difficult for the above models to accomplish a performance-computation trade-off. In this paper, we propose a Dual-former that exploits CNNs-based local feature extraction in the encoder and decoder parts. The self-attention operation only performs global modeling capability in the latent layer, which attracts SOTA performance in multiple tasks and has less computational complexity than previous methods.

\begin{figure*}[!t]

    \centering
    \includegraphics[width=16cm]{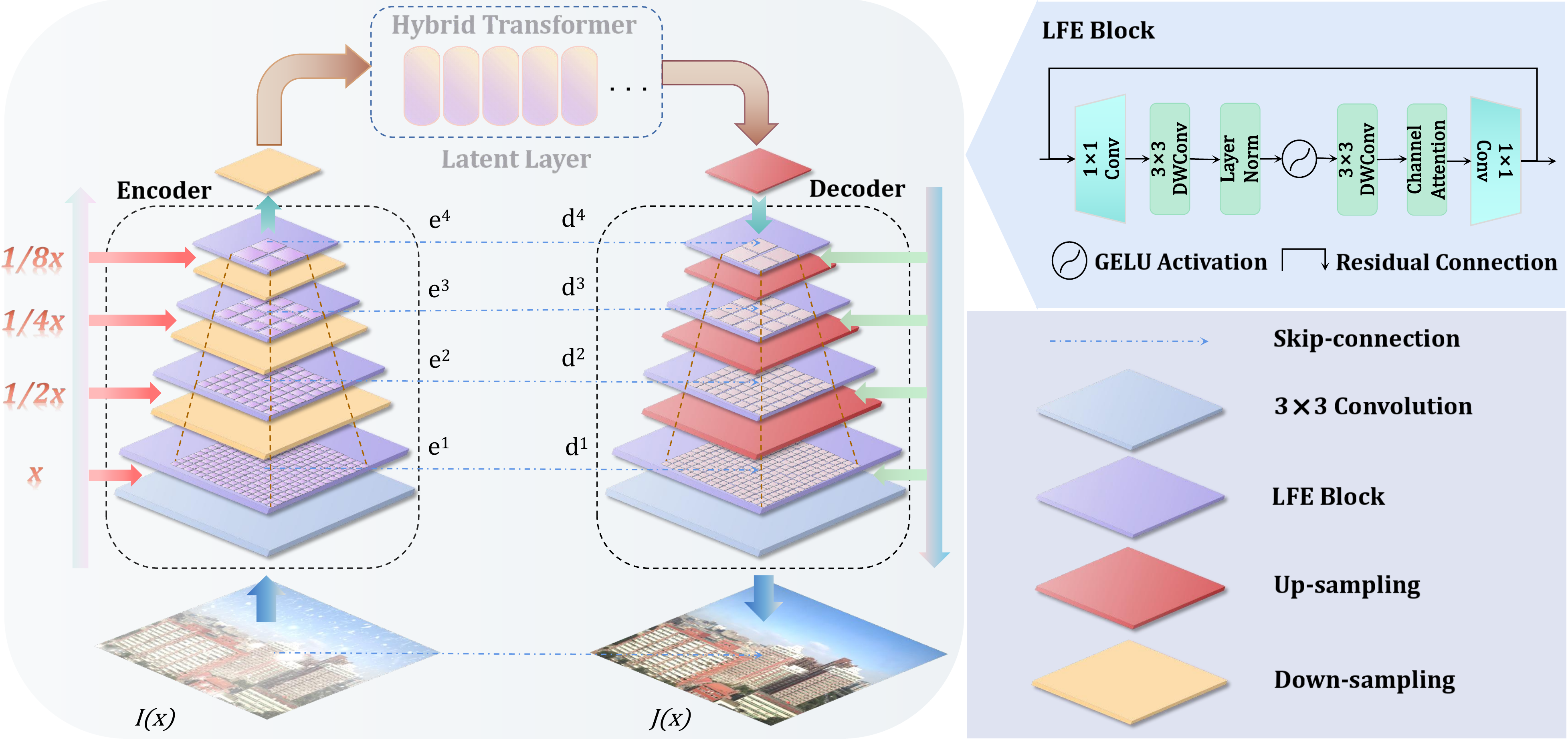}
    \caption{The left part is the proposed Dual-former framework. For the encoder-decoder, it consists of CNNs-based Local Feature Extraction (LFE) module. We devise the Hybrid Transformer Block (HTB) and embed it into the latent layer of the network. The upper right corner is the specific composition of our LFE block.}
    \label{overview}
\end{figure*}
\vspace{-0.5cm}
%-----------------------------------Our method
\section{Our Method}
\subsection{Overall Architecture of Dual-former}
The overall architecture of Dual-former is shown in Fig.\ref{overview}. Dual-former employs the CNNs-based encoder and decoder. The Hybrid Transformer block is only embedded in the latent layer to obtain the trade-off between performance and computational complexity. Given an input image $I(x)\in \mathbb{R}^{ H\times W\times 3}$, we first utilize a 3$\times$3 convolution to project the image into feature. Then we adopt the proposed Local Feature Extraction (LFE) block to capture the local information, and use a 3$\times$3 convolution aiming to reduce the image's resolution while expanding channel dimensions. Such action enriches the local receptive field in each encoder stage, which plays a crucial role in image restoration. The encoder can be formulated as follows:
\begin{equation}
    X_{0}^{e} = \operatorname{LFE}(\operatorname{\varphi_{3}}(I(x))),
\end{equation}
\begin{equation}
X_{i+1}^{e} = \operatorname{LFE}(X_{{i}}^{e}\in \mathbb{R}^{ \frac{H}{2^{i}}\times \frac{W}{2^{i}}\times C_{i}}\downarrow),
\end{equation}
where $X_{i+1}^{e} \in \mathbb{R}^{ \frac{H}{2^{i+1}}\times \frac{W}{2^{i+1}}\times C_{i+1}}$ is the ($i$+1)-th feature of encoder, H, W, and C respectively denote the height, width and channel number of input image $I(x)$. $X$ is the feature of the input image. $\varphi_{3}$ and $\operatorname{LFE}$ are the 3$\times$3 convolution and LFE module. $\downarrow$ is the down-sampling operation. $i,i$+1 and $e$ represent the $i$-th and ($i$+1)-th stage of encoder. We set the i to 0, 1, 2 and 3 for each layer in this paper. 

Unlike the previous ViT-based image restoration architectures that utilize the Transformer block in the entire network~\cite{liang2021swinir,lee2022knn,wang2022uformer}, we only use the self-attention mechanism in the latent layer. Due to its progressively extensive local receptive field, global self-attention could leverage local information well to perform global modeling. In Hybrid Transformer Block (HTB), we introduce self-attention from different dimension perspectives, where we perform spatial global modeling while paying attention to the channel self-information. In addition, we offer an Adaptive Control Module (ACM) to fusion the channel-spatial self-information. The Multi-branch Feed Forward Network is proposed to further improve the capacity of local inductive bias compared with previous feed-forward designs. In the latent layer, we can express it as the following formulas:

\begin{equation}
X_{4}^{l} = \operatorname{HTB}(X_{4}^{e}\in \mathbb{R}^{ \frac{H}{16}\times \frac{W}{16}\times C_{l}}),
\end{equation}
where the $l$ denotes the latent layer and $C_{l}$ is set to 384 in our paper. For the decoder part, we continue to compose the configuration of each stage with LFE module and gradually up-sample to the original resolution. Corresponding to each encoder stage, we add a skip connection to the decoder to avoid information loss. The decoder part can be defined by:
\begin{equation}
    X_{i-1}^{d} = \operatorname{LFE}(X_{{i}}^{d}\in \mathbb{R}^{ \frac{H}{2^{i}}\times \frac{W}{2^{i}}\times C_{i}}\uparrow + X^{e}_{i-1}),
\end{equation}
    
\begin{equation}
    J(x) = \operatorname{\varphi_{3}}(X_{0}^{d} + I(x)),
\end{equation}
where $X_{4}^{d}=X_{4}^{l}$, ${X_{i-1}^{d}}\in \mathbb{R}^{ \frac{H}{2^{i-1}}\times \frac{W}{2^{i-1}}\times C_{i-1}}$, the $d$ represents the decoder part of network. $J(x)$ is the clean image restored from degraded version. $\uparrow$ is the up-sampling operation. $+$ denotes skip connection that we utilize between each encoder and decoder~\cite{zamir2021restormer}. We set the i to 4, 3, 2 and 1 for each decoder. 

\begin{figure*}[!t]
    \centering
    \includegraphics[width=16cm]{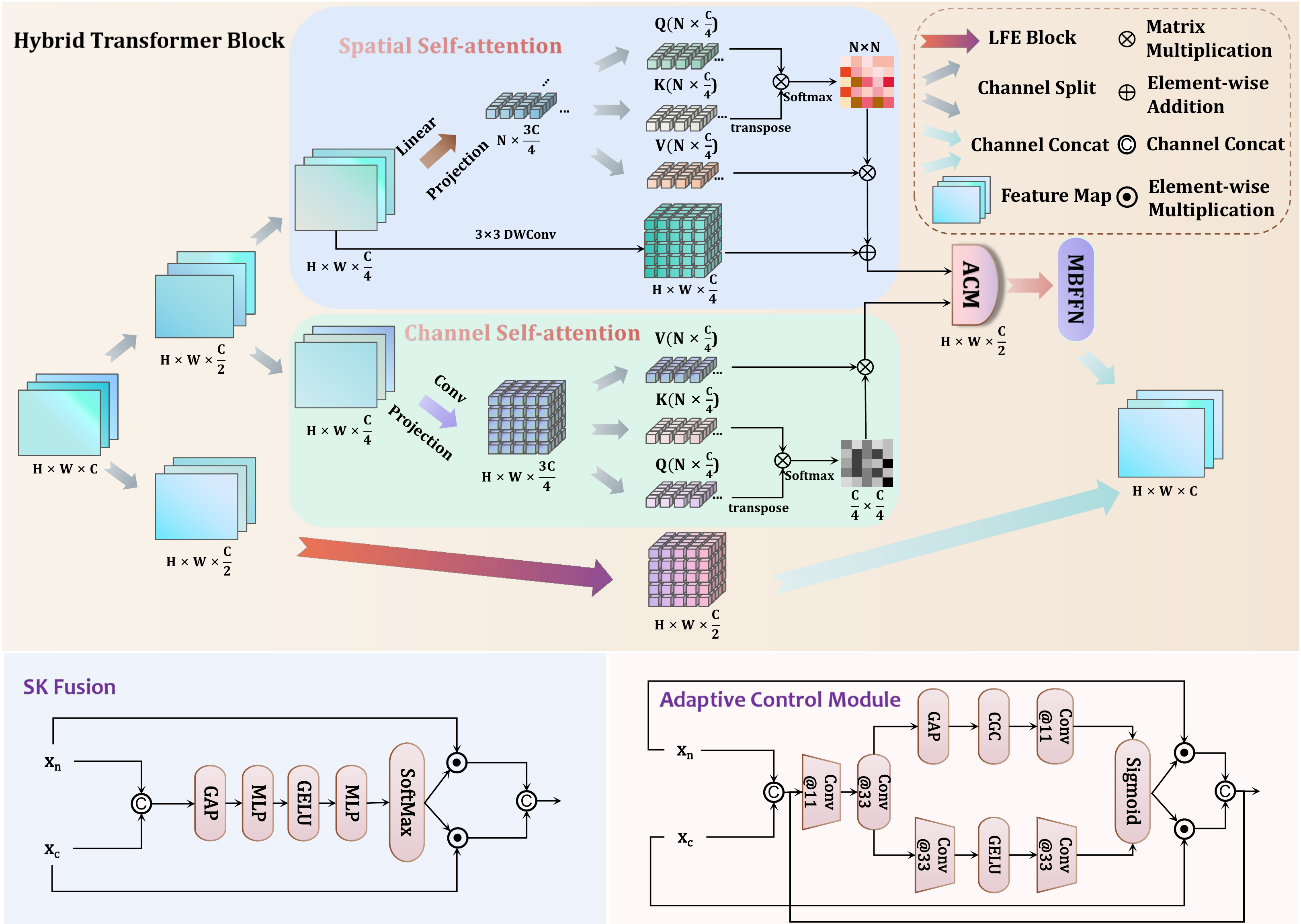}
    \caption{The upper part is our Hybrid Transformer Block (HTB). It consists of a mixture of spatial self-attention and channel self-attention mechanisms, as well as LFE modules. The lower left corner is the operation of SK fusion~\cite{song2022vision}. The lower right corner is the Adaptive Control Module (ACM) we designed, which strengthens the fusion of spatial interaction information and channel interaction information.}
    \label{HTB}
\end{figure*}

%-----------------------------------------------Local Feature Extraction Block

\subsection{Local Feature Extraction Block}
For image restoration, local feature modeling determines the details of the restored image~\cite{zamir2021restormer}. Convolution performs the merit of local modeling because of local inductive bias. Compared with the vanilla self-attention operation~\cite{vit}, the computation of convolution is not sensitive to the resolution of feature maps due to its $O(n)$ complexity. As shown in Fig.\ref{overview}, we present the Local Feature Extraction block (LFE) in the encoder and decoder to obtain rich feature information while saving computational complexity. Specifically, we believe high dimension space is crucial for the feature to perform local information modeling. Different from the ~\cite{zamir2021restormer,resnet}, we first adopt a 1$\times$1 convolution to expand the number of dimensions. Then to achieve the trade-off between performance and computation complexity, we use two 3$\times$3 depth-wise convolutions to capture local information instead of 3$\times$3 convolutions. Besides, we adopt a Layer Normalization operation and a GELU activation function between them to consolidate training and optimize the nonlinear ability of the network. Channel attention has been proven effective for image restoration~\cite{chen2022simple}. However, its performance is limited in shallow layers due to the number of channels. To tackle this issue, we add channel attention~\cite{hu2018squeeze} after convolutions to further improve channel information interaction in high-dimension space. Finally, we utilize a residual connection~\cite{resnet} to enhance the proposed block's robustness.

% For image restoration, detailed information from the high-resolution degradation image is imperative. As shown in Fig.(\ref{overview}), in the encoder-decoder part, we devise the Local Feature Extract block (LFE), which aims to obtain rich local information for image restoration. Benefiting from up and down-sample convolution, our overall backbone exhibits a pyramid structure. This design is to enlarge the receptive field for the LFE block in each stage and avoid the substantial computational complexity instead of the quadratic complexity of token number for the vision transformer. 
% In the LFE module, we only focus on local modeling information for the degraded image. Inspired by MobileNet-V2~\cite{sandler2018mobilenetv2}, we follow the Inverted Residual Bottleneck, projecting the features to high-dimensional space to improve the representation capability. We employ two depth-wise convolutions in the high-dimensional layer to extract local information. In addition, we adopt a Layer Normalization operation and a GELU activation function to consolidate training and optimize the nonlinear ability of our model. It is worth mentioning that we add a channel attention mechanism to interact information between channels at the end, which can make up for the lack of depth-wise convolution in channel dimension and enhance the overall performance. Finally, we utilize a residual connection~\cite{resnet} to improve the robustness of the architecture.

\subsection{Hybrid Transformer Block}
In the latent layer, we design a Hybrid Transformer Block (HTB) to exploit the merit of it for global degraded modeling from channel and spatial dimensional perspectives, while the convolution block ensures the extraction of local information. The proposed HTB can be expressed as:
\begin{equation}
    \operatorname{HTB}(X^{l}) = [\operatorname{HS}\&\operatorname{ACM}(X^{g}), \operatorname{LFE}(X^{p})],
\end{equation}
where $[\cdot]$ denotes the channel-wise concatentation operation. $\operatorname{HS}$, $\operatorname{ACM}$ and $\operatorname{LFE}$ represent the Hybrid Self-attention, Adaptive Control Module and Local Feature Extraction. $g$ and $p$ denote that feature will be modeled globally and locally. 
\subsubsection{Hybrid Self-attention with Adaptive Control Module}
%\subsubsubsection{11111}
The vanilla vision transformer~\cite{vit} consists of a self-attention and a feed-forward network:
\begin{equation}
\begin{aligned}
&\hat{X} = X + \operatorname{Self-attention}(X),\\
&\hat{Y} = \hat{X} + \operatorname{FFN}(\hat{X}),
\end{aligned}
\end{equation}
For an input feature $X\in \mathbb{R}^{ H\times W\times C}$, it reshapes feature into a three-dimension vector sequence $X\in \mathbb{R}^{ N \times C}$. Before the self-attention operation, it usually learns $W_{q}\in \mathbb{R}^{ C \times C}$, $W_{k}\in \mathbb{R}^{ C \times C}$, $W_{v}\in \mathbb{R}^{ C \times C}$ to project $X$ into $\mathbf{Q}$uery ($XW_{q}\in \mathbb{R}^{ N \times C}$), $\mathbf{K}$ey ($XW_{k}\in \mathbb{R}^{ N \times C}$) and $\mathbf{V}$alue ($XW_{v}\in \mathbb{R}^{ N \times C}$). Then utilize the dot product between them and the $\operatorname{Softmax}$ operation to form a self-attention matrix, which is used to interact own information. Finally,  the value performs the global modeling by the attention matrix. The general self-attention can be defined as follows:
\begin{equation}
    X' = \operatorname{Softmax}(\frac{\mathbf{QK^{\mathrm{T}}}}{\sqrt{C}}) \times \mathbf{V}.
\end{equation}

Following the above self-attention of vanilla transformer in low-level tasks~\cite{lee2022knn,Guo_2022_CVPR,valanarasu2022transweather,wang2022uformer}, we note that the action explores the long-range modeling between patches to calculate the $N \times N$ attention map but lacks the global modeling from the channel dimension, whose uneven distributions determine the channel global modeling is detrimental to image restoration. Therefore, to achieve non-trivial performance and lower complexity, we premeditate the entire information flow and devise a hybrid self-attention module in the latent layer. Specifically, as shown in Fig.\ref{overview}, given a input feature $X^{l}\in \mathbb{R}^{ H \times W \times C}$, we split it along the channel dimension into $X^{g}\in \mathbb{R}^{ H \times W \times \frac{C}{2}}$ and $X^{p}\in \mathbb{R}^{ H \times W \times \frac{C}{2}}$. Then we use $X^{g}$ feature to perform global modeling. (Another feature $X^{p}$ is processed by LFE module in parallel. See the (3) paragraph: Parallel Local Feature Extract Module).
Next, we split the feature $X^{g}\in \mathbb{R}^{ H\times W \times \frac{C}{2}}$ along the channel dimension into the $X_{c}\in \mathbb{R}^{ H \times W\times\frac{C}{4}}$ and $X_{s}\in \mathbb{R}^{ H \times W\times\frac{C}{4}}$ to process the channel self-attention (CSA) and spatial self-attention (SSA). We respectively employ a linear and convolution layer to offer the learnable parameters, which projects $X_{c}$ and $X_{s}$ into Query ($\mathbf{Q}$), Key ($\mathbf{K}$), and Value ($\mathbf{V}$) that will perform different dimensional perspectives self-attention:
\begin{equation}
\begin{aligned}
&\mathbf{Q_{c}}^{ N \times \frac{C}{4}} = {X_{c}W_{cq}},  \mathbf{K_{c}}^{ N \times \frac{C}{4}} = {X_{c}W_{ck}},  \mathbf{V_{c}}^{ N \times \frac{C}{4}} = {X_{c}W_{cv}},\\
&\mathbf{Q_{s}}^{ N \times \frac{C}{4}} = {X_{s} W_{sq}},  \mathbf{K_{s}}^{ N \times \frac{C}{4}} = {X_{s}W_{sk}},  \mathbf{V_{s}}^{ N \times \frac{C}{4}} = {X_{s}W_{sv}},
\end{aligned}
\end{equation}
where the subscript c and s represent the parameters that will perform CSA and SSA.
Inspired by~\cite{zamir2021restormer}, we attach the depth-wise convolution after the convolution projection to keep the local contextual information in the channel self-attention part. After obtained three crucial $Q_{\left\{c,s\right\}}$ $K_{\left\{c,s\right\}}$ and $V_{\left\{c,s\right\}}$, we can use the dot product operation to obtain the self-information matrix in the channel dimension, which focus on channel interaction:
\begin{equation}
    A_{c}\in \mathbb{R}^{ \frac{C}{4} \times \frac{C}{4}} = \operatorname{Softmax}(\frac{\mathbf{Q_{c}^\mathrm{T}K_{c}}}{\sqrt{\frac{C}{4}}}).
\end{equation}
Similarly, we can get the self-attention map of the interaction between patches at spatial level:
\begin{equation}
    A_{s}\in \mathbb{R}^{ N \times N} = \operatorname{Softmax}(\frac{\mathbf{Q_{s}K_{s}^{\mathrm{T}}}}{\sqrt{\frac{C}{4}}}). 
\end{equation}
Then we use matrix multiplication for the self-attention relationship map $A_{\left\{c,s\right\}}$ and the original image to process feature with different levels $X'_{c}\in \mathbb{R}^{ N \times \frac{C}{4}}$, $X'_{s}\in \mathbb{R}^{ N \times \frac{C}{4}}$. In addition, we utilize a 3$\times$3 depth-wise convolution to improve the local inductive bias in spatial self-attention. For the above operations, we are inspired by the multi-head self-attention mechanism~\cite{vit} (MSA) to split the dimension into multiple heads. For instance, $\mathbf{Q} = [\mathbf{Q}_{1};\mathbf{Q}_{2};\mathbf{Q}_{3};...;]$ where $\mathbf{Q}_{i}\in \mathbb{R}^{ N \times \frac{C}{4H}}$ represents the i-th head of total H number heads. The $\mathbf{Q}_{i}$ implements the self-attention mechanism with other multi-head $\mathbf{K}_{i}$ and $\mathbf{V}_{i}$ respectively, and we concatenate the multi-output in the last, allowing the model to learn complex information. 

Overall, we can express channel self-attention and spatial self-attention separately as follows:
\begin{equation}
\begin{aligned}
&X'_{c} = \operatorname{MSA}(\mathbf{V_{c}} \times \operatorname{Softmax}(\frac{\mathbf{Q_{c}^\mathrm{T}K_{c}}}{\sqrt{\frac{C}{4}}})) ,\\
&X'_{s} = \operatorname{MSA}(\operatorname{Softmax}(\frac{\mathbf{Q_{s}K_{s}^{\mathrm{T}}}}{\sqrt{\frac{C}{4}}}) \times \mathbf{V_{s}}) + \operatorname{DWConv}(X_{s}),
\end{aligned}
\end{equation}
For the image restoration, hybrid self-attention that gives multiple dimensional perspectives information outperforms single interaction (In Table \ref{sa ab}).

Obtained the interaction feature of different dimensional levels, how to guide and control each other between them is a crucial problem we are concerned about. Inspired by~\cite{li2019selective}, \cite{song2022vision} propose SK fusion layer to fuse the multiple branches information, as shown in Fig.\ref{HTB}. However, we notice that it only adopts channel dimension attention to control information flow, which brings limited improvement compared with concatenate operation. In this paper, we present $\mathbf{A}$daptive $\mathbf{C}$ontrol $\mathbf{M}$odule (ACM). It considers the various concerns of the channel and spatial self-attention, which can better perform information fusion and solve the above drawback superiorly. 
We first stitch together two different information streams and exploit the 1$\times$1 and 3$\times$3 convolutions to mine the general feature information $X_{o}$. For the feature processed by channel self-attention $X'_{c}$, we utilize the spatial level attention to offset the shortcoming in spatial dimension adaptively:
\begin{equation}
    X'_{c} = X'_{c} \times \operatorname{Sigmoid}(\operatorname{\varphi_{3}}(\sigma(\operatorname{\varphi_{3}}(X_{o})))) ^ { H \times W \times 1},
\end{equation}
where $\operatorname{\sigma}$ is the GELU activation and $\operatorname{Sigmoid}$ is the Sigmoid function, we utilize convolution to reduce the dimension number gradually. Correspondingly, we use Global AvgPooling to notice the channel interaction for the feature $X'_{s}$:
\begin{equation}
        X'_{s} = X'_{s} \times \operatorname{Sigmoid}(\operatorname{CGC}(\operatorname{\mathcal{GAP}}(X_{o}))) ^ { 1 \times 1 \times \frac{C}{4}},
\end{equation}
wherein the $\operatorname{CGC}$ includes (Conv-GELU-Conv). Finally, we concatenate the $X'_{c}$ and $X'_{s}$
along with the channel dimension, and add the residual connection to input the feed-forward network. Experiments demonstrate that our ACM performs impressively (In Tabel \ref{acm ab}).

%----------------------------------------------------------------Multi-branch Feed Forward Network

\begin{figure*}[!t]

    \vspace{-1mm}
    \begin{center}
        \begin{tabular}{cccccccc}
        
\includegraphics[width = 0.12\linewidth]{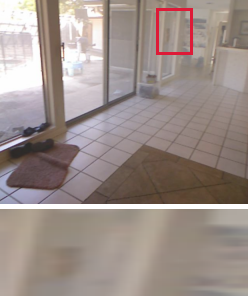} &\hspace{-4.5mm}

\includegraphics[width = 0.12\linewidth]{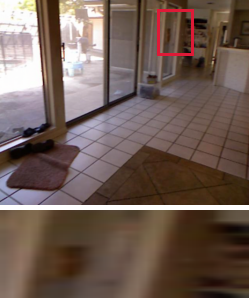} &\hspace{-4.5mm}
\includegraphics[width = 0.12\linewidth]{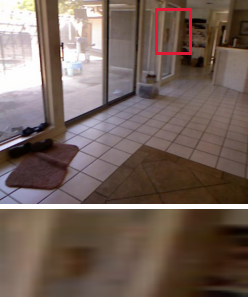} &\hspace{-4.5mm}
\includegraphics[width = 0.12\linewidth]{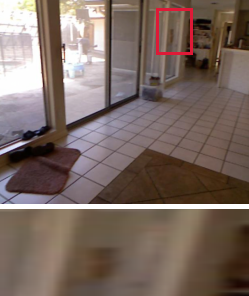} &\hspace{-4.5mm}

\includegraphics[width = 0.12\linewidth]{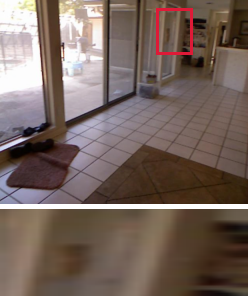} &\hspace{-4.5mm}
\includegraphics[width = 0.12\linewidth]{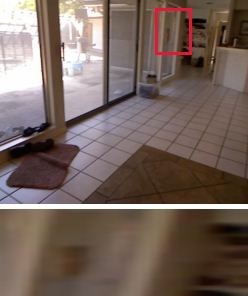} &\hspace{-4.5mm}
\includegraphics[width = 0.12\linewidth]{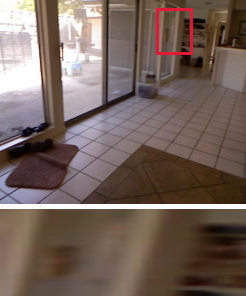} &\hspace{-4.5mm}
\\
\includegraphics[width = 0.12\linewidth]{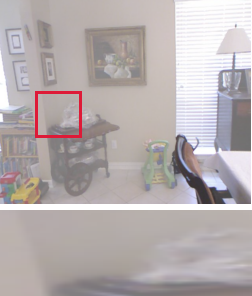} &\hspace{-4.5mm}

\includegraphics[width = 0.12\linewidth]{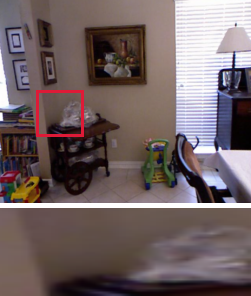} &\hspace{-4.5mm}
\includegraphics[width = 0.12\linewidth]{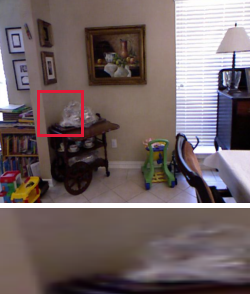} &\hspace{-4.5mm}
\includegraphics[width = 0.12\linewidth]{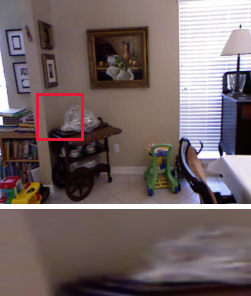} &\hspace{-4.5mm}

\includegraphics[width = 0.12\linewidth]{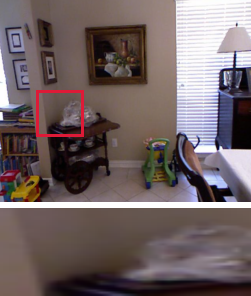} &\hspace{-4.5mm}
\includegraphics[width = 0.12\linewidth]{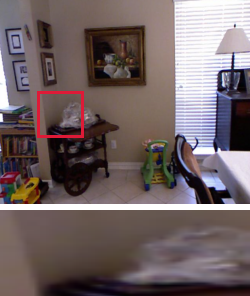} &\hspace{-4.5mm}
\includegraphics[width = 0.12\linewidth]{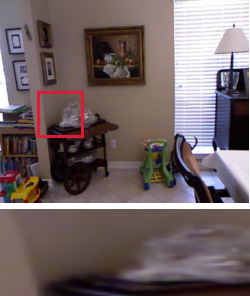} &\hspace{-4.5mm}

\\
\small{(a)Input}  &\hspace{-4mm} \small{(b)MSBDN~\cite{msbdn1}} &\hspace{-4mm} \small{(c)FFA-Net~\cite{qin2020ffa}} &\hspace{-4mm} \small{(d)Dehamer~\cite{Guo_2022_CVPR}}&\hspace{-4mm} \small{(e)MAXIM~\cite{tu2022maxim}}&\hspace{-4mm} \small{(f)Dual-former} &\small{(g)Ground-truth} 
\end{tabular}
\end{center}
\caption{Visual comparisons of various deraining methods on SOTS-Indoor~\cite{li2018benchmarking} dataset. Please zoom in for a better illustration.}\label{fig.haze}
\end{figure*}

\subsubsection{Multi-branch Feed Forward Network}
\begin{figure}[!t]
\setlength{\belowcaptionskip}{-0.5cm}%调整caption与下文的距离
    \centering
    \includegraphics[width=8cm]{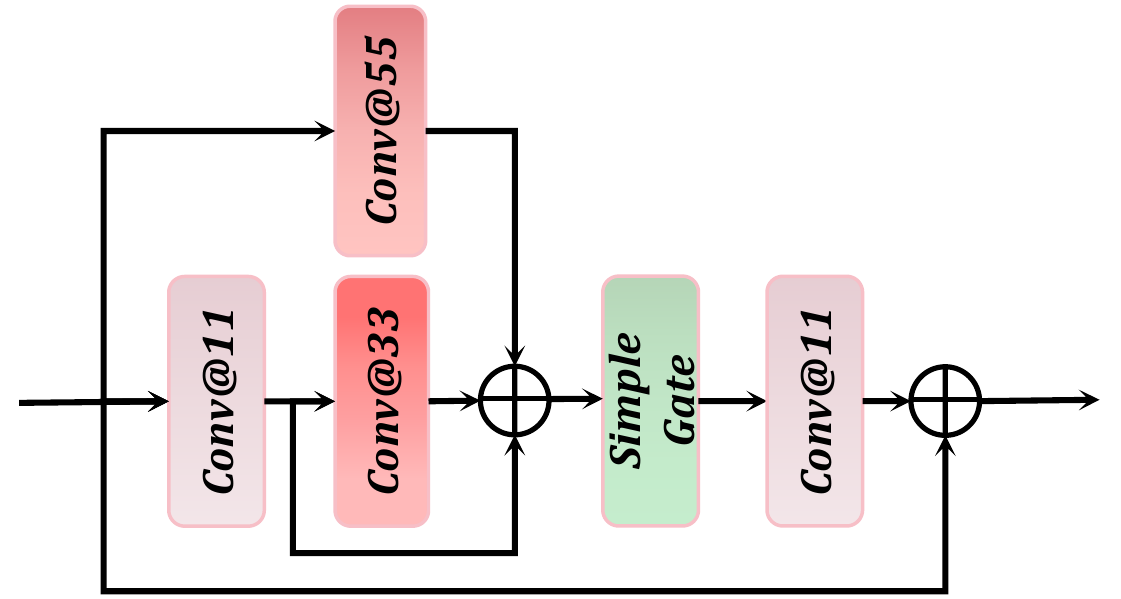}
    \caption{Architecture of Multi-branch Feed-forward Network.}
    \label{mbffn}
\end{figure}

As a vital part of the transformer block, the feed-forward network enriches the representation performance of the model. Previous architectures consist of multilayer perceptron (MLP), which limits the ability of the network to leverage the locally~\cite{swim,vit}. ConvFFN~\cite{lee2022knn,PVT} applied depth-wise convolution to extract local contextual information. However, we note that in image restoration tasks, multi-scale information can better improve model performance due to its diverse receptive fields. Consequently, we propose the Multi-branch Feed-forward Network (MBFFN), which offers the different kernels of convolution to capture multi-size contextual information. As depicted in Fig.\ref{mbffn}, we first adopt 1$\times$1 convolution to expand the feature channels (we set expand factor $\beta$ = 2 to save computation in this paper). Next, we employ the multi-branch to aggregate different scale information by 3$\times$3 and 5$\times$5 convolution. Inspired by ~\cite{chen2022simple}, we use the Simple Gate (SG) operation to replace the GELU activation before the final 1$\times$1 convolution. Specifically, given an input $X\in \mathbb{R}^{ H \times W \times C}$, our MBFFN can be formulated as:
\begin{equation}
\begin{aligned}
&Y = \operatorname{\varphi_{1}}(\operatorname{SG}(\operatorname{MB}(\operatorname{\varphi_{1}}(X)))) + X,\\
&\operatorname{MB}(X) = \operatorname{\varphi_{3}}(X) + \operatorname{\varphi_{5}}(X) + X,
\end{aligned}
\end{equation}
where the $Y$ is the output feature.

%------------------------------------Parallel Local Feature Extract Module
\subsubsection{Parallel Local Feature Extract Module}
In the HTB, besides the self-attention mechanism and feed-forward network, we also design a local feature extraction module in parallel with them along the channel dimension. It provides the ability of local information acquisition in high-dimensional space, which is complementary to the global modeling ability of the parallel hybrid self-attention mechanism. It is worth mentioning that we increase the highest number of dimensions due to the parallelism of the convolution blocks. However, compared with blocks composed of all transformers, our hybrid transformer block takes into account the balance between performance and computation, laying the foundation for the efficiency of the overall network. 

%--------------------------------------Loss Function
\subsection{Loss Function}
To facilitate our model for better image restoration performance. we utilize PSNR loss~\cite{chen2021hinet} as our reconstruction loss function. It can be expressed as follows:
\begin{equation}
        \left.L_{p s n r}=-\operatorname{PSNR}(I(X)), Y\right),
\end{equation}
where PNSR($\cdot$) denotes the PSNR metrics, X and Y are the input and its corresponding ground-truth image, $I$($\cdot$) represents our Dual-former. In addition, we also use perceptual loss, which provides supervision at the perceptual perspective level:
\begin{equation}
L_{perceptual }=\sum_{j=1}^{2} \frac{1}{C_{j} H_{j} W_{j}} \left\| \phi_{j}(\mathcal{I}(X)\right)-\phi_{j}\left(Y)\right\|_{1},
\end{equation}
wherein the $C_{j}$, $H_{j}$, $W_{j}$ are the number of channels, length and width of the feature map extracted in the j-th hidden feature of VGG19~\cite{simonyan2014very} model, respectively. $\phi_{j}$ means the specified j-th layer of the network.

Therefore, our overall loss function can be expressed as follows:
\begin{equation}
    L = \lambda_{1} L_{psnr} + \lambda_{2} L_{perceptual },
\end{equation}
where $\lambda_{1}$ and $\lambda_{2}$ are set to 1 and 0.2 in our paper.

\begin{table*}[!t]
\centering
\caption{Deraining results on five datasets.  Bold and underline indicate the best and the second best metrics. }\label{rainresults}
\resizebox{16cm}{!}{
\renewcommand\arraystretch{1.1}
\begin{tabular}{l|cccccccccccc|cc}
\toprule[1.2pt]

\multirow{2}*{Method} & \multicolumn{2}{c}{ Test100~\cite{zhang2019image} } & \multicolumn{2}{c}{ Rain100H~\cite{yang2017deep} } & \multicolumn{2}{c}{ Rain100L~\cite{yang2017deep} } & \multicolumn{2}{c}{$\text { Test2800 ~\cite{fu2017removing} }$} & \multicolumn{2}{c}{$\text { Test1200~\cite{DIDMDN} }$} & \multicolumn{2}{||c|}{$\text { Average }$}& {\multirow{2}{*}{\#GFLOPs}} & {\multirow{2}{*}{\#Param}} \\\cline{2-13}
 & PSNR $\uparrow$ & SSIM $\uparrow$ & PSNR $\uparrow$ & SSIM $\uparrow$ & PSNR $\uparrow$ & SSIM $\uparrow$ & $\text { PSNR }$ $\uparrow$ & $\text { SSIM }$ $\uparrow$ & $\text { PSNR }$ $\uparrow$ & $\text { SSIM }$ $\uparrow$ & \multicolumn{1}{||c}{$\text { PSNR }$ $\uparrow$} & \multicolumn{1}{c|}{$\text { SSIM }$ $\uparrow$} \\\hline
(TIP'17)DerainNet~\cite{fu2017derainnet} & 22.77 & 0.810 & 14.92 & 0.592 & 27.03 & 0.884 & 24.31 & 0.861 & 23.38 & 0.835 &  \multicolumn{1}{||c}{22.48} & 0.796 &${88.287}$G & 0.75M\\
(CVPR'18)DIDMDN~\cite{DIDMDN} & 22.56 & 0.818 & 17.35 & 0.524 & 25.23 & 0.741 & 28.13 & 0.867 & 29.65 & 0.901 & \multicolumn{1}{||c}{24.58} & 0.770   &$7.31G$ & 0.37K\\
(CVPR'19)UMRL~\cite{UMRL} & 24.41 & 0.829 & 26.01 & 0.832 & 29.18 & 0.923 & 29.97 & 0.905 & 30.55 & 0.910 & \multicolumn{1}{||c}{28.02} & 0.880 & 16.50G & 0.98K\\
(ECCV'18)RESCAN~\cite{rescan} & 25.00 & 0.835 & 26.36 & 0.786 & 29.80 & 0.881 & 31.29 & 0.904 & 30.51 & 0.882 & \multicolumn{1}{||c}{28.59} & 0.857 &${32.32}$G& 0.15M\\
(CVPR'19)PreNet~\cite{prenet} & 24.81 & 0.851 & 26.77 & 0.858 & {32.44} & {0.950} & 31.29 & 0.904 & 30.51 & 0.882 & \multicolumn{1}{||c}{29.42} & 0.897  &${66.58}$G& 0.17M\\
(CVPR'20)MSPFN~\cite{mspfn} & {27.50} & {0.876} & {28.66} & {0.860} & 32.40 & 0.933 & 31.75 & 0.916 & 31.36 & 0.911 & \multicolumn{1}{||c}{30.75} & 0.903 &{708.44}G& 21.00M \\
(CVPR'21)MPRNet~\cite{mpr} & \underline{30.27} & {0.897} & {30.41}& {0.890} & 36.40 & 0.965 & {33.64} & {0.938} & \underline{32.91} & 0.916 & \multicolumn{1}{||c}{32.73} & {0.921} &148.55G & 3.64M \\
(CVPR'22)KiT~\cite{lee2022knn} & {30.26} & \underline{0.904} & \underline{30.47} & \underline{0.897} & \underline{36.65} & \underline{0.969} & \underline{33.85} & \underline{0.941} & {32.81} & \underline{0.918} & \multicolumn{1}{||c}{\underline{32.81}} & \underline{0.926}& 43.08G &20.60M\\\hline\hline
\gr Dual-former & $\mathbf{30.63}$ & $\mathbf{0.908}$ & $\mathbf{30.85}$ &${0.894}$ & $\mathbf{36.99}$ & $\mathbf{0.969}$ & ${33.51}$ & ${0.937}$ & {32.66} & $\mathbf{0.919}$ & \multicolumn{1}{||c}{$\mathbf{32.93}$} & $\mathbf{0.926}$ & 9.28G & 14.23M\\
\bottomrule[1.2pt]
\end{tabular}}
\end{table*}

\begin{figure*}[!t]

    \vspace{-1mm}
    \begin{center}
        \begin{tabular}{ccccccccc}
        
\includegraphics[width = 0.11\linewidth]{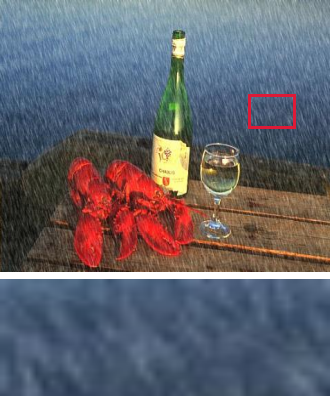} &\hspace{-4.5mm}

\includegraphics[width = 0.11\linewidth]{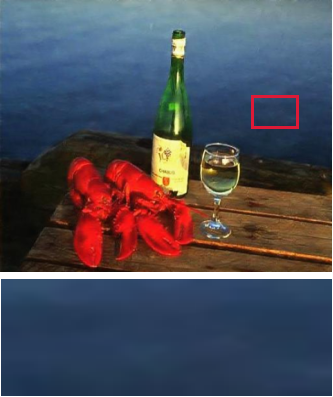} &\hspace{-4.5mm}

\includegraphics[width = 0.11\linewidth]{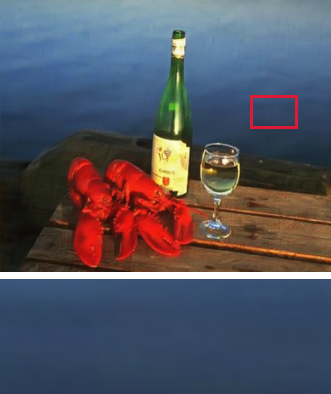} &\hspace{-4.5mm}

\includegraphics[width = 0.11\linewidth]{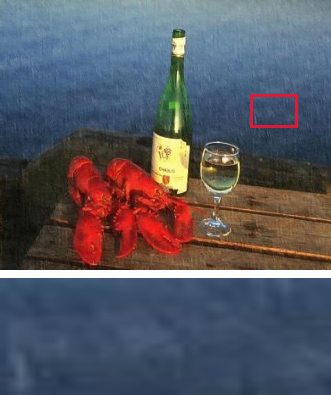} &\hspace{-4.5mm}
\includegraphics[width = 0.11\linewidth]{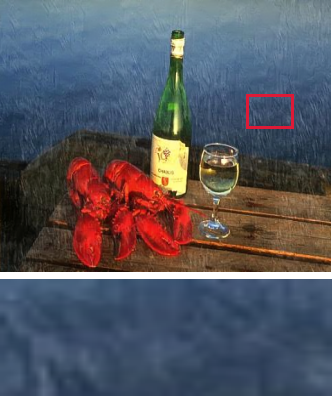} &\hspace{-4.5mm}
\includegraphics[width = 0.11\linewidth]{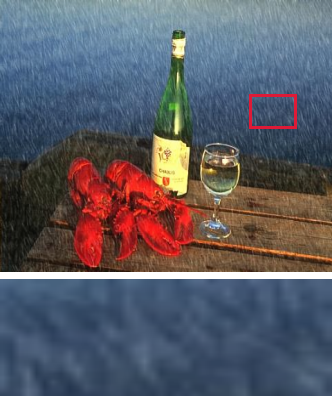} &\hspace{-4.5mm}
\includegraphics[width = 0.11\linewidth]{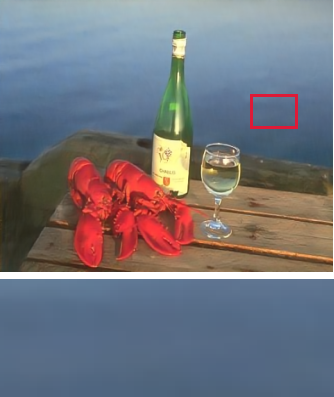} &\hspace{-4.5mm}
\includegraphics[width = 0.11\linewidth]{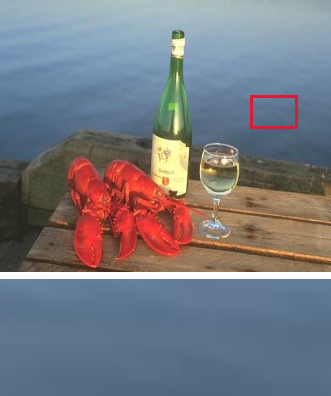} &\hspace{-4.5mm}
\\
\includegraphics[width = 0.11\linewidth]{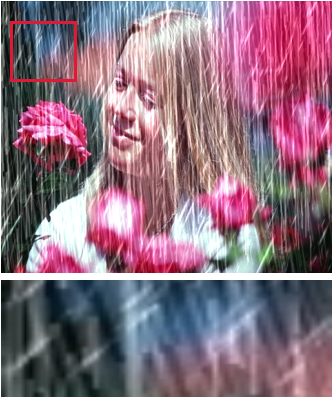} &\hspace{-4.5mm}

\includegraphics[width = 0.11\linewidth]{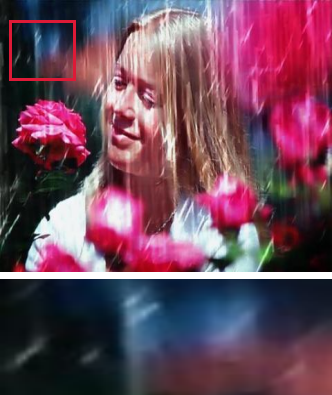} &\hspace{-4.5mm}

\includegraphics[width = 0.11\linewidth]{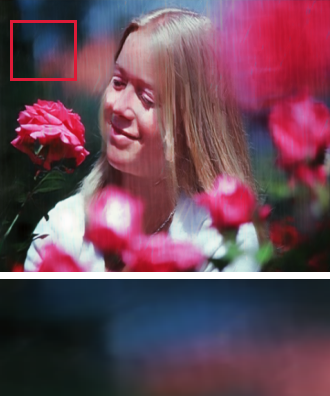} &\hspace{-4.5mm}
\includegraphics[width = 0.11\linewidth]{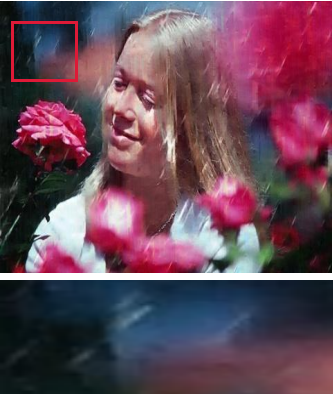} &\hspace{-4.5mm}
\includegraphics[width = 0.11\linewidth]{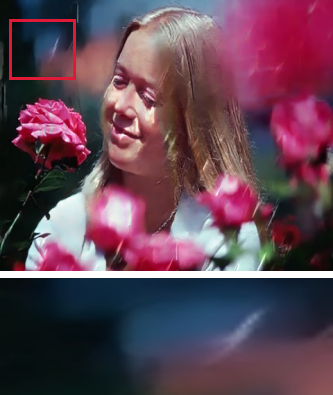} &\hspace{-4.5mm}
\includegraphics[width = 0.11\linewidth]{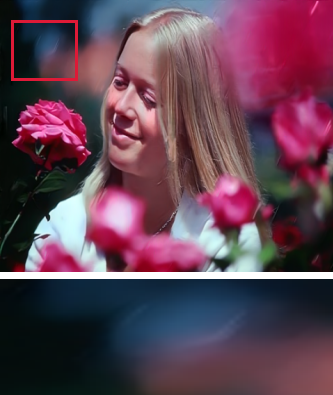} &\hspace{-4.5mm}
\includegraphics[width = 0.11\linewidth]{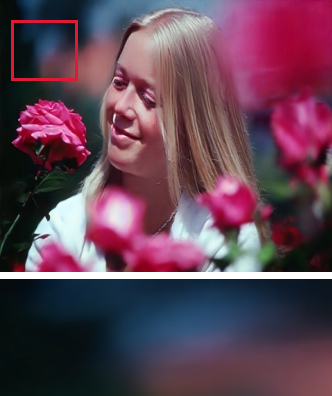} &\hspace{-4.5mm}
\includegraphics[width = 0.11\linewidth]{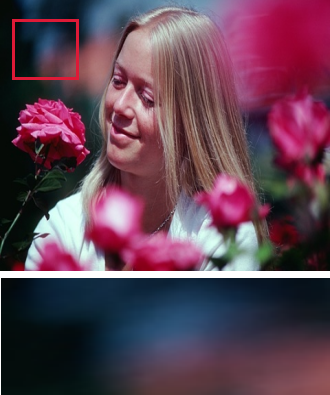} &\hspace{-4.5mm}

\\

\scriptsize{(a)Input}  &\hspace{-4mm} \scriptsize{(b)DIDMDN~\cite{DIDMDN}}  &\hspace{-4mm} \scriptsize{(c)UMRL~\cite{UMRL}}&\hspace{-4mm} \scriptsize{(d)RESCAN~\cite{rescan}}&\hspace{-4mm} \scriptsize{(e)PreNet~\cite{prenet}} &\hspace{-4mm} \scriptsize{(f)MPRNet~\cite{mpr}}&\hspace{-4mm} \scriptsize{(g)Dual-former} &\hspace{-4mm} \scriptsize{(h)Ground-truth} 
\end{tabular}
\end{center}
\caption{Visual comparisons of various deraining methods on Test100~\cite{zhang2019image} dataset. Please zoom it for a better illustration.}\label{fig.derain}
\end{figure*}
%------------------------Experiments
\section{Experiments}
%------------------------Implementation Details
\subsection{Implementation Details}
We train our model with the Adam optimizer, whose initial momentum is set to 0.9 and 0.999. We set the initial learning rate to be 0.0002 with cycle learning rate decay. We present 256$\times$256 patch size from the original image after data augmentation as our input and train them for 800 epochs for each low-level visual image restoration task. In terms of data augmentation, we flip horizontally and randomly rotate the image to a fixed angle. We choose the 1-th and 3-th layers of VGG19~\cite{simonyan2014very} to adopt perceptual loss.

%-----------------------Architecture
\vspace{-0.5cm}
\subsection{Architecture Design}
For a concise description, we introduce the specific configurations of our Dual-former. We gradually expand the dimension of our encoder and set it to $\{$28, 32, 64, 128$\}$. We increase the number of LFE blocks for each encoder stage to $\{$4, 5, 7, 8$\}$. In the LFE block, we keep the original dimensions at the shallowest level and utilize the 1$\times$1 convolution to double the dimension numbers in other stages. In the latent layer, we increase the dimension numbers to 384 and stack 14 Hybrid Transformer Blocks to improve the network's performance, whose number of multi-head is set to 8. At the same time, we follow the ~\cite{zamir2021restormer} to apply the skip connection between the encoder and decoder. 
%-----------------------Tasks and Metrics
\vspace{-0.5cm}
\subsection{Tasks and Metrics}
Image restoration for severe weather is a challenging task (i.e., haze, rain, snow). Most previous models attract remarkable results in a single adverse weather restoration task. However, the performance in other tasks is not satisfactory. In this paper, we conduct extensive experiments on three adverse weather tasks and a total of ten datasets (Haze4K~\cite{dmt}, SOTS~\cite{li2018benchmarking}, Test100~\cite{zhang2019image}, Rain100H~\cite{yang2017deep}, Rain100L~\cite{yang2017deep}, Test2800~\cite{fu2017derainnet}, Test1200~\cite{DIDMDN}, CSD~\cite{hdcwnet}, SRRS~\cite{chen2020jstasr} and Snow100K~\cite{liu2018desnownet}) to verify the excellent property of Dual-former. For evaluation metrics, we adopt PSNR and SSIM~\cite{wang2004image} to be consistent with previous methods.

\vspace{-0.5cm}

%-----------------------Image Dehazing
\subsection{Image Dehazing}
Image dehazing is a typical low-level vision task in image restoration for degraded weather. we compare the performance of Dual-former with other SOTA dehazing approaches: DCP~\cite{dcp}, NLD~\cite{nld}, DehazeNet~\cite{dehazenet}, GDN~\cite{gdn}, MSBDN~\cite{msbdn1}, DA~\cite{da}, FFA-Net~\cite{qin2020ffa}, DMT-Net~\cite{dmtnet}, Dehamer~\cite{Guo_2022_CVPR} and MAXIM~\cite{tu2022maxim}. We train our Dual-former on Haze4K~\cite{dmt} dataset and ITS~\cite{li2018benchmarking}, the Haze4K testset~\cite{dmt} and SOTS-Indoor~\cite{li2018benchmarking} as evaluation datasets to validate the restored performance.

Table \ref{hazeresults} shows that our Dual-former achieves significant and unusual performance while keeping efficient characteristics due to its low GFLOPs compared with the state-of-the-art models. For the latest method (Dehamer and MAXIM), we exceed quantitative results by 1.91dB on SOTS-Indoor datasets compared with MAXIM, only using 4.2$\%$ computation \textbf{(216.00G $\rightarrow$ 9.28G)}. In addition, we outperform the Dehamer substantially while adopting 10.7$\%$ parameters \textbf{(132.45M $\rightarrow$ 14.23M)}. Fig.\ref{fig.haze} illustrates that the image produced by our design is superior in recovering details and overall effect, which is sharper and cleaner in local features than other state-of-the-art algorithms. To emphasize the impressive performance of image dehazing, we also present the visual comparison of real-world images in Fig.\ref{realsnow}. We catch that Dual-former attracts the non-trivial effect of removing the haze veiling from the degraded version. On the contrary, other SOTA approaches such as MAXIM and Dehamer, almost fail to recover the clean images.

\begin{table}[!h]
\centering

\caption{Dehazing results on the Haze4K~\cite{dmtnet} and SOTS-Indoor~\cite{li2018benchmarking} datasets.  Bold and underline indicate the best and the second best metrics. }\label{hazeresults}
\resizebox{8.5cm}{!}{
\renewcommand\arraystretch{1.1}

\begin{tabular}{l|cc|cc|cc}
\toprule[1.2pt]

{\multirow{2}*{Method}} & \multicolumn{2}{c|}{ Haze4k~\cite{dmtnet} } & \multicolumn{2}{c|}{ SOTS Indoor~\cite{li2018benchmarking} } & {\multirow{2}*{\#GFLOPs}} & {\multirow{2}*{\#Param}} \\
 &PSNR $\uparrow$ & SSIM $\uparrow$ & PSNR $\uparrow$ & SSIM $\uparrow$ \\
\hline (TPAMI'10)DCP~\cite{dcp} & $14.01$ & $0.76$ & $15.09$ & $0.76$ &$\mathbf{-}$ & $\mathbf{-}$\\
(CVPR'16)NLD~\cite{nld} & $15.27$ & $0.67$ & $17.27$ & $0.75$ &$\mathbf{-}$ & $\mathbf{-}$\\
(TIP'16)DehazeNet~\cite{dehazenet} & $19.12$ & $0.84$ & $20.64$ & $0.80$ & {0.52G} & 0.01M\\
(ICCV'17)GDN~\cite{gdn} & $23.29$ & $0.93$ & $32.16$ & ${0.98}$ &{21.49G} & 0.96M\\
(CVPR'20)MSBDN~\cite{msbdn1} & $22.99$ & $0.85$ & $33.79$ & ${0.98}$ &{41.58G} & 31.35M\\
(AAAI'20)FFA-Net~\cite{qin2020ffa} & $26.96$ & $0.95$ & $36.39$ & $0.98$ &{288.34G}& 4.6M \\
(ACMMM'21)DMT-Net~\cite{dmtnet} & $\underline{28.53}$ & $\underline{0.96}$ & $\mathbf{-}$ & $\mathbf{-}$ &{80.71G} & 54.9M\\
(CVPR'22)Dehamer~\cite{Guo_2022_CVPR} & $\mathbf{-}$ & $\mathbf{-}$ & {\underline{36.63}} & \underline{0.99} &{59.15G} & 132.45M\\
(CVPR'22 Oral)MAXIM~\cite{tu2022maxim} & $\mathbf{-}$ & $\mathbf{-}$ & {\underline{38.11}} & \underline{0.99} &{216.00G} & 14.10M\\
\hline\hline \gr Dual-former & $\mathbf{32.85}$ & $\mathbf{0.98}$ & ${\mathbf{40.02}}$ & $\mathbf{0.99}$ &9.28G&14.23M\\ 
\bottomrule[1.2pt]
\end{tabular}}
\end{table}

%-----------------------Image Deraining
\subsection{Image Deraining}
For the image deraining task, we follow the settings of most current mainstream methods~\cite{zamir2021restormer}, applying our Dual-former to train on multiple datasets, which include 13712 paired images. In the test stage, we evaluate the performance of our framework on five datasets, containing Rain100H~\cite{yang2017deep}, Rain100L~\cite{yang2017deep}, Test100~\cite{zhang2019image}, Test2800~\cite{fu2017removing}, and Test1200~\cite{DIDMDN}. We utilize PSNR and SSIM as our reference metrics and convert the image to Ycrcb color space to compute metrics for the Y channel, which is consistent with previous SOTA methods. 

We present the compared results of the previous latest method in Table \ref{rainresults}. It's noticed that most SOTA strategies for rain streak removal need massive computation to do more processing, which aims to cover the difference of five datasets such as MPRNet \textbf{(148.55G)} and KiT \textbf{(43.08G)}. We outperform the latest approaches while only keeping the computation in the low overhead \textbf{(9.28G)}. 
Fig.(\ref{fig.derain}) is a visual comparison of other methods for the deraining task. We observe that our Dual-former can remove small rain streaks thoroughly instead of retaining a vestige of rain spots, especially in dense rain images where MPRNet almost fails to remove rain streaks. For the restored details, our approach is closer to the ground-truth. In a word, our manner has impressive visual effects compared with state-of-the-art methods.
Furthermore, the deraining performance on the real-world dataset is also shown in Fig.\ref{realsnow}. We notice that the previous methods such as RESCAN\cite{rescan} and MPRNet\cite{mpr}, remnant the large rain streaks in the restored image. Though PReNet\cite{prenet} and MSPFN\cite{mspfn} can remove the lots of streaks, they are still hard to handle the residual rain spots perfectly. The image processed by Our Dual-former keeps the details while eliminating the rain degradations, which demonstrates that the proposed algorithm obtains dramatic performance on the real-world dataset.
%--------------------------------gflops-parameters

\begin{table}[!h]
\centering

\caption{Quantitative comparison of various desnowing approaches on the CSD~\cite{hdcwnet}, SRRS~\cite{chen2020jstasr} and Snow 100K~\cite{liu2018desnownet} datasets.  Bold and underline indicate the best and the second best metrics. }\label{snowresults}
\resizebox{8.5cm}{!}{
\renewcommand\arraystretch{1.1}
\begin{tabular}{l|cc|cc|cc|cc|cc}
\toprule[1.2pt]
\multirow{2}*{Method}& \multicolumn{2}{c|}{ CSD (2000) } & \multicolumn{2}{c|}{ SRRS (2000) } & \multicolumn{2}{c|}{ Snow 100K (2000) } & \multicolumn{2}{c}{\multirow{2}*{\#GFLOPs}}& \multicolumn{2}{c}{\multirow{2}*{\#Param}}\\\cline{2-7}
& PSNR $\uparrow$ & SSIM $\uparrow$ & PSNR $\uparrow$ & SSIM $\uparrow$ & PSNR $\uparrow$ & SSIM $\uparrow$\\
\hline
% Zheng  & 20.21 &0.79 & 19.21 & 0.80 &&\\
% Eigen & 18.63 & 0.66 & 17.36 & 0.66 \\
(TIP'18)DesnowNet~\cite{liu2018desnownet} & 20.13 & 0.81 &20.38 &0.84& 30.50 & 0.94 & \multicolumn{2}{c}{1.7KG} & \multicolumn{2}{c}{15.6M}\\
(CVPR'18)CycleGAN~\cite{engin2018cycle}& 20.98 & 0.80 &20.21 &0.74& 26.81 & 0.89 & \multicolumn{2}{c}{42.38G} & \multicolumn{2}{c}{7.84M}\\
(CVPR'20)All in One~\cite{allinone} &26.31 &{0.87} &24.98 &0.88& 26.07&0.88 & \multicolumn{2}{c}{12.26G} &\multicolumn{2}{c}{44M}\\
(ECCV'20)JSTASR~\cite{chen2020jstasr} &27.96 &0.88 & 25.82 & 0.89 & 23.12 & 0.86 & \multicolumn{2}{c}{-}& \multicolumn{2}{c}{65M}\\
(ICCV'21)HDCW-Net~\cite{hdcwnet} & {29.06} &{0.91} &{27.78} &{0.92} & \underline{31.54} &\underline{0.95} & \multicolumn{2}{c}{9.78G} & \multicolumn{2}{c}{6.99M}\\
(CVPR'22)TransWeather~\cite{valanarasu2022transweather} &$\underline{31.76}$ &$\underline{0.93}$ &$\underline{28.29}$  &$\underline{0.92}$ &$\underline{31.82}$ &${0.93}$ & \multicolumn{2}{c}{5.64G} & \multicolumn{2}{c}{21.9M}\\
\hline\hline 
\gr  Dual-former &$\mathbf{35.90}$&$\mathbf{0.97}$ & $\mathbf{\mathbf{32.92}}$ &$\mathbf{0.96}$& $\mathbf{34.99}$ & $\mathbf{0.97}$ & \multicolumn{2}{c}{9.28G} & \multicolumn{2}{c}{14.23M}\\
\bottomrule[1.2pt]
\end{tabular}}
\end{table}

\begin{figure*}[!t]
    \vspace{-1mm}
    \begin{center}
        \begin{tabular}{ccccccc}
\includegraphics[width = 0.14\linewidth]{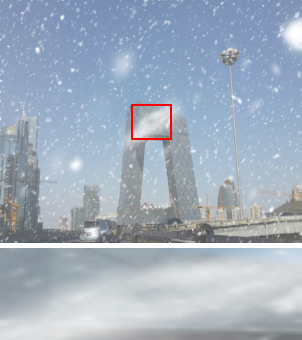} &\hspace{-3.5mm}
\includegraphics[width = 0.14\linewidth]{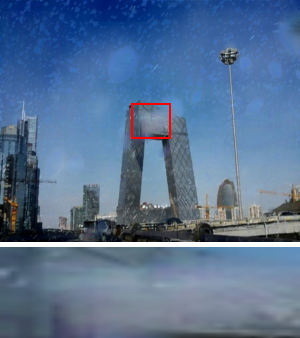} &\hspace{-3.5mm}
\includegraphics[width = 0.14\linewidth]{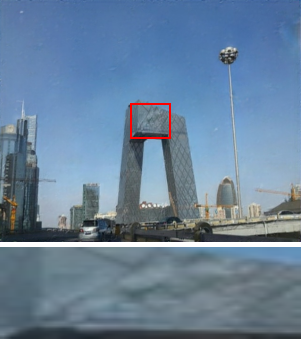} &\hspace{-3.5mm}
\includegraphics[width = 0.14\linewidth]{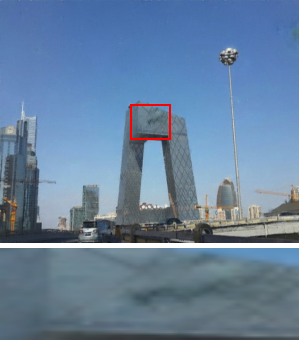} &\hspace{-3.5mm}
\includegraphics[width = 0.14\linewidth]{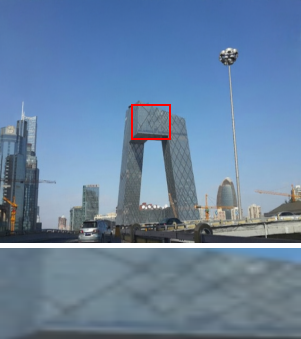} &\hspace{-3.5mm}
\includegraphics[width = 0.14\linewidth]{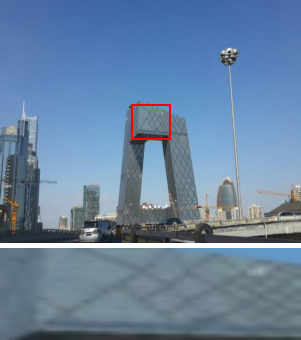} & \hspace{-3.5mm}
%&\vspace{-4mm}
\\
\includegraphics[width = 0.14\linewidth]{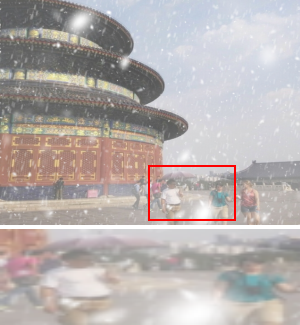} &\hspace{-3.5mm}
\includegraphics[width = 0.14\linewidth]{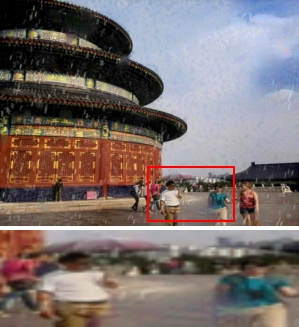} &\hspace{-3.5mm}
\includegraphics[width = 0.14\linewidth]{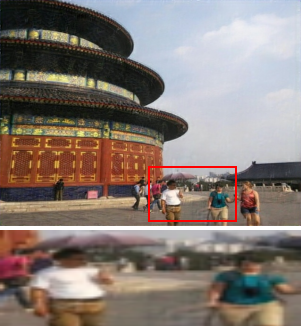} &\hspace{-3.5mm}
\includegraphics[width = 0.14\linewidth]{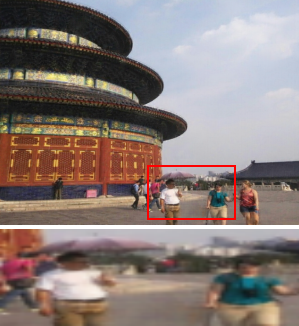} &\hspace{-3.5mm}
\includegraphics[width = 0.14\linewidth]{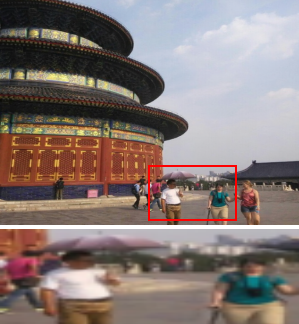} &\hspace{-3.5mm}
\includegraphics[width = 0.14\linewidth]{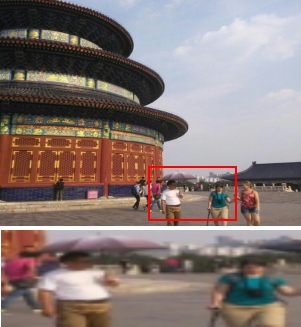} & \hspace{-3.5mm}
\\
\footnotesize{(a)Input}  &\hspace{-3.5mm} 
\footnotesize{(b)JSTASR~\cite{chen2020jstasr}} &\hspace{-3.5mm} 
\footnotesize{(c)HDCW-Net~\cite{hdcwnet}} &\hspace{-3.5mm}  
\footnotesize{(d)TransWeather~\cite{valanarasu2022transweather}} &\hspace{-3.5mm}
\footnotesize{(e)Dual-former(Ours)} &\hspace{-3.5mm}
\footnotesize{(f)Ground-truth} &\hspace{-3.5mm}\\
        \end{tabular}
    \end{center}
\caption{Visual comparison of our Dual-former and the state-of-the-art methods on the CSD~\cite{hdcwnet} synthetic dataset . Please zoom it to watch the details better.}\label{fig:snowcomparison}
\end{figure*}
%-----------------------Image Desnowing

\begin{figure*}[!h]
    \centering
    \includegraphics[width=16cm]{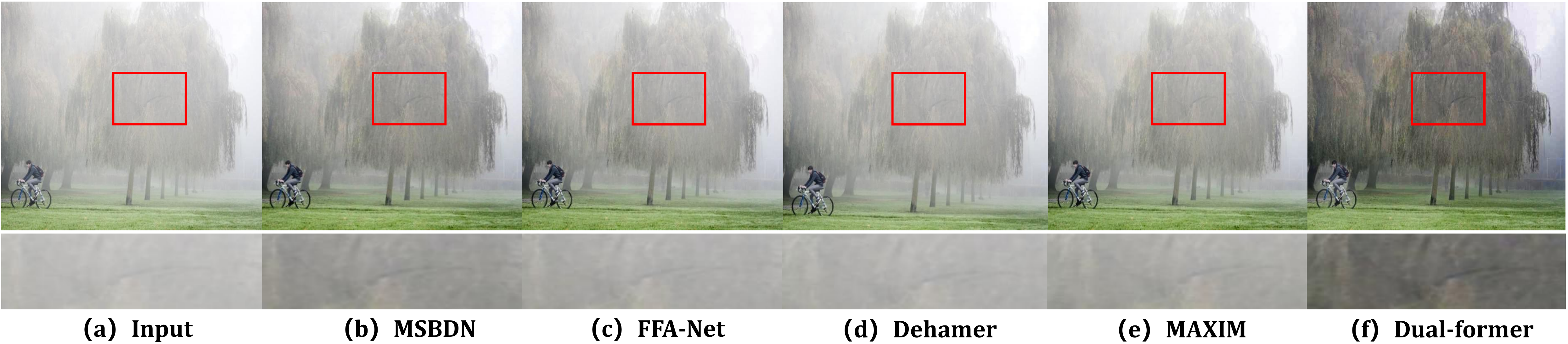}
    \label{realhazy}
\end{figure*}
\vspace{-0.5cm}
\begin{figure*}[!h]
    \centering
    \includegraphics[width=16cm]{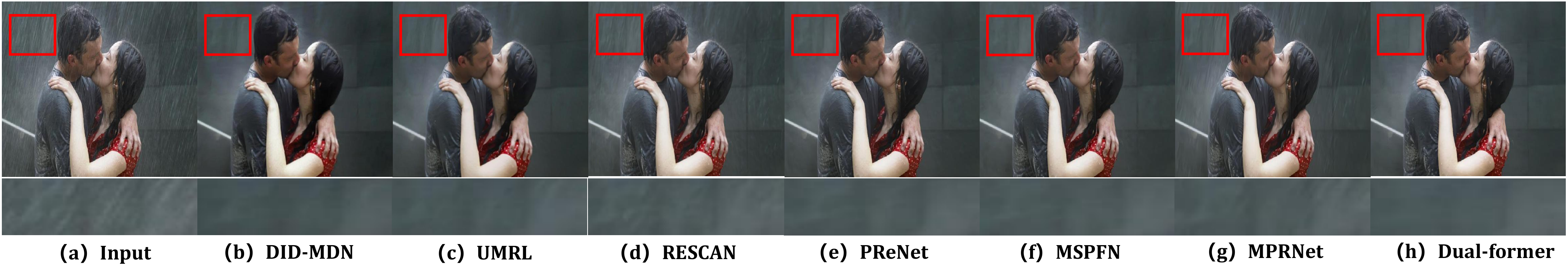}
    \label{realrain}
\end{figure*}
\begin{figure*}[!h]
    \centering
    \includegraphics[width=16cm]{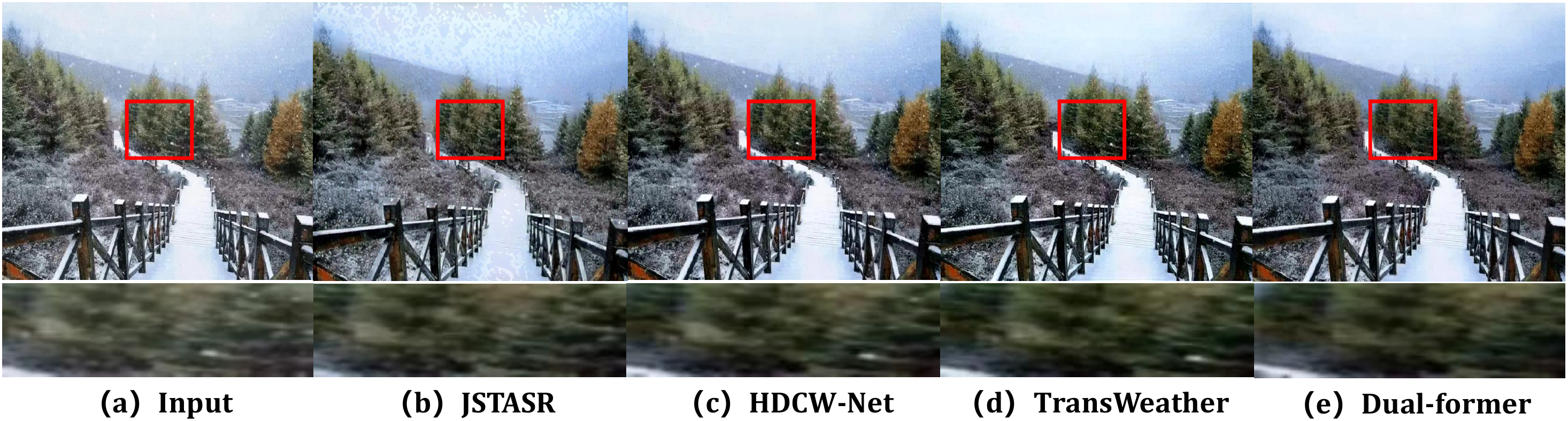}
    \caption{Visual comparison between adverse weather removal methods on real-world datasets (dehazing, deraining, desnowing). Please zoom them to watch the details better.}
    \label{realsnow}
\end{figure*}
\subsection{Image Desnowing}
For single image snow removal, we train and test our Dual-former on three snow datasets, CSD~\cite{hdcwnet} Snow100K~\cite{liu2018desnownet}, SRRS~\cite{chen2020jstasr}. We follow the latest single image desnowing method~\cite{hdcwnet} for training and testing on different testsets to ensure reasonableness and fairness (sample 2000 images from testset for each dataset). For the models without training on desnowing datasets, we re-train them and choose the best result for comparison. 

The results of previous methods are shown in Table \ref{snowresults}. We notice that Dual-former shows substantial growth performance compared with state-of-the-art methods on CSD~\cite{hdcwnet} datasets \textbf{(31.76dB $\rightarrow$ 35.90dB)}, the same as SRRS~\cite{chen2020jstasr} and Snow100K~\cite{liu2018desnownet}. This excellent performance proves that our method is highly competitive in desnowing task. We also carry out the visual comparison, the effects are presented in Fig.\ref{fig:snowcomparison}. From Fig.\ref{fig:snowcomparison}, we found that our snow removal effect is better than the previous SOTA methods to remove various snow degradation and restore image color. At the same time, our Dual-former can better distinguish snow marks and object details while the HDCW-Net~\cite{hdcwnet} and Transweather~\cite{valanarasu2022transweather} almost recover failure. Furthermore, our features in the detailed enlarged image are more stunning and close to the ground-truth. We also compare the desnowing performance of SOTA manners on the real-world image, and the results are shown in Fig.\ref{realsnow}. We observe that previous methods JSTASR\cite{chen2020jstasr}, HDCW-Net\cite{hdcwnet} and TransWeather\cite{valanarasu2022transweather} are difficult to resolve the small degradations, such as snow spots. However, Dual-former tackles this problem and has the best visual effect compared with the latest approaches.

%---------------------------------Image Desnowing

\subsection{Effective Receptive Field Analysis}
\begin{figure}[!h]
\setlength{\belowcaptionskip}{-1cm}%调整caption与下文的距离
    \centering
    \includegraphics[width=8.5cm]{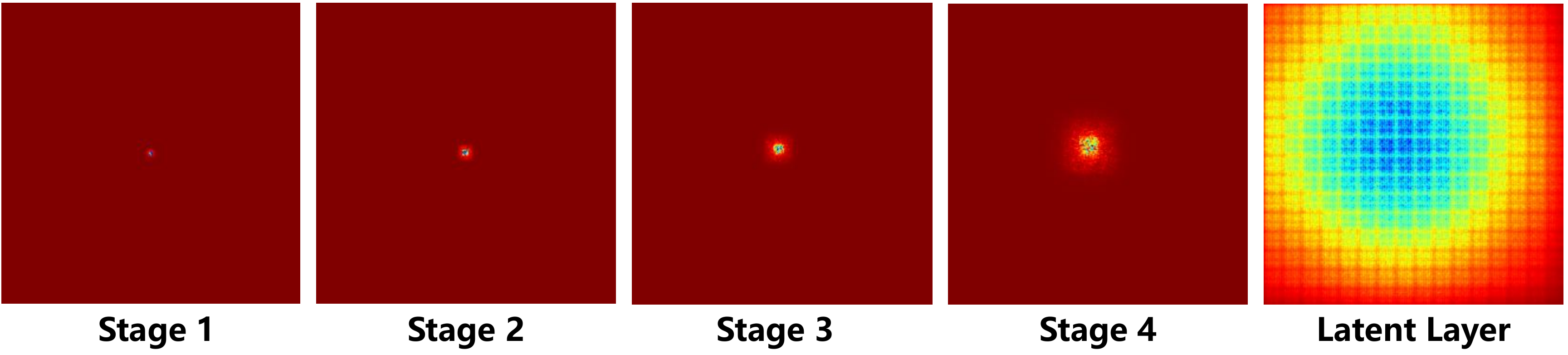}
    \caption{Effective Receptive Field of different stages on Rain100H~\cite{yang2017deep} testset (average over 100 images).}
    \label{receptive}
\end{figure}

Our main core design idea uses CNNs in the encoder part to extract local features, and embeds self-attention into latent to enhance the global modeling ability. Therefore, we employ the effective receptive (ERF) field~\cite{luo2016understanding} of different stages in the proposed architecture. From Fig.\ref{receptive}, such ERF demonstrates that the LFE module in the encoder focuses on the local information extraction, while expanding the ERF progressively. In the latent layer, the hybrid transformer block performs global modeling primarily, which has a large ERF. Such a design fully considers the previous conclusions~\cite{raghu2021vision} while significantly reducing the computational overhead.
 
\vspace{-0.5cm}

\subsection{Parameters and Computational complexity Analysis} 
In this section, we explore the advantages of Dual-former in parameters and computation. Parameters and FLOPs are within our scope of the designed architecture. Nevertheless, achieving impressive performance while keeping suitable parameters and computational overheads is challenging. In Table (\ref{hazeresults}-\ref{rainresults}), We present the comparison results of the latest approaches in the different recovering tasks. We notice that our model's computation is considerably less than most state-of-the-art methods. For image dehazing, we only use 4.2$\%$ computation complexity to surpass the MAXIM~\cite{tu2022maxim}'s performance.
Furthermore, the amount of parameters is much less than most previous methods. It is worth mentioning that the parameter of FFA-Net~\cite{qin2020ffa} is 4.6M, while it needs a vast computational burden of 288.34G (256$\times256$). Also, in the rain removal task, it is difficult for MPRNet~\cite{mpr} to balance the lightweight design while being a small computation. 

% To highlight our strengths, we also test the model speed of 512$\times$512 resolution on RTX3090 GPU, shown in Table \ref{speed1}. We can observe that our model outperforms previous methods while having significantly less computational complexity than previous SOTA methods. For the high-resolution processing speed, we attract the best performance despite their low number of parameters. On the task of removing snow, although TransWeather~\cite{valanarasu2022transweather} is slightly less computational overhead than us, our Dual-former dramatically improves performance with fewer parameters, which proves that our method can exhibit a parameter-performance trade-off.

\begin{figure*}[!h]
    \centering
    \includegraphics[width=16cm]{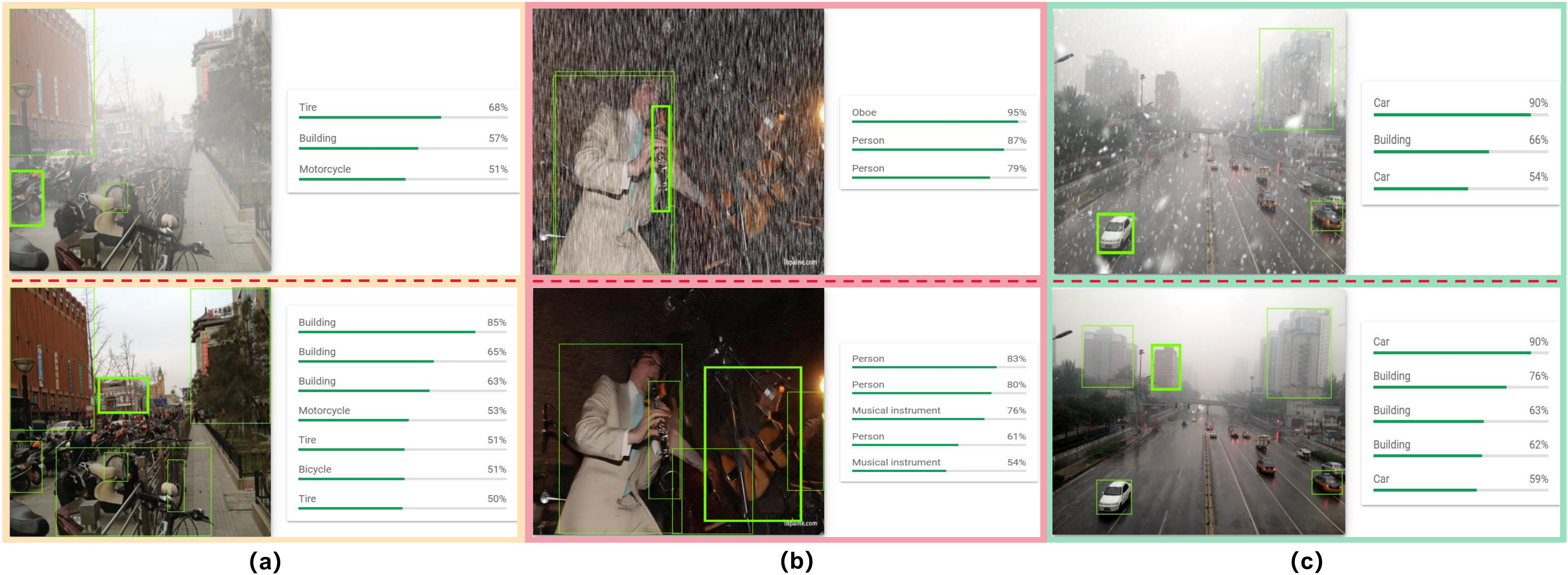}
    \caption{(a), (b) and (c) are the quantifiable performance for high-level vision tasks. The upper contents are the detection confidences of degraded images (haze, rain, snow) and the lower contents are the detection confidences of corresponding restored versions from Dual-former.}
    \label{highlevel}
\end{figure*}

% \begin{table}[!h]
% \centering
% %\setlength{\abovecaptionskip}{0.7cm}
% \caption{512×512 resolution speed comparison of different models (/s). Underline indicate the best metrics.}\label{speed1}
% \resizebox{8cm}{!}{
% \renewcommand\arraystretch{1.1}
% \begin{tabular}{ccccccc}
% \toprule[1.2pt]
% \textbf{Model} & \multicolumn{2}{c}{\textbf{\#Param}} & \multicolumn{2}{c}{\textbf{\#GFLOPs}} & \textbf{Speed}   \\
% \midrule[0.15em]
% FFA-Net~\cite{qin2020ffa}& \multicolumn{2}{c}{4.6M} & \multicolumn{2}{c}{1153.34G} & 0.17s  \\
%  Dehamer~\cite{Guo_2022_CVPR}& \multicolumn{2}{c}{132.45M} & \multicolumn{2}{c}{235.77G} & 0.05s  \\
%  MPRNet~\cite{mpr} &  \multicolumn{2}{c}{3.64M} & \multicolumn{2}{c}{594.21G} & 0.09s &  \\
%  Dual-former(Ours)&  \multicolumn{2}{c}{14.23M} & \multicolumn{2}{c}{31.48G} & \underline{0.04s}  \\
% \bottomrule[1.2pt]
% \end{tabular}}
% \end{table}
\vspace{-0.3cm}
\subsection{The merits of Dual-former}
\begin{figure}[!h] %通栏
\begin{minipage}[t]{0.5\linewidth} %调节两个子图左右间距
\centering
\includegraphics[width=1\textwidth]{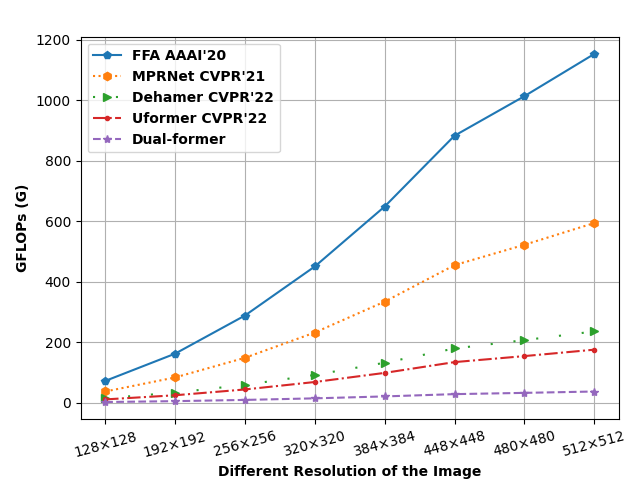} %调节单个子图大小
\label{gflops} %引用标签
\end{minipage}%
\begin{minipage}[t]{0.5\linewidth}
\centering
\includegraphics[width=1\textwidth]{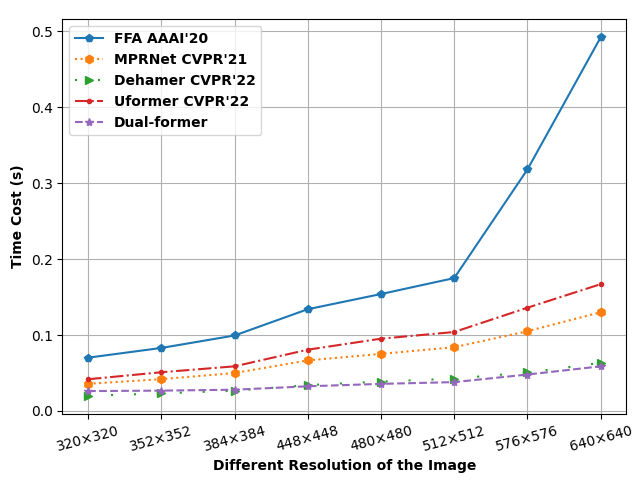}
\label{speed}
\end{minipage}%
\vspace{-0.5mm}
    \caption{Left: \textbf{GFLOPs of our method and SOTA methods with respect to different image resolutions.} Our Dual-former has fewer GFLOPs than all competitors, and it enjoys the lowest GFLOPs growth rate inspired by input image size. Right: \textbf{Time cost corresponding to various input image sizes of proposed Dual-former and SOTA approaches.} Dual-former is superior in time cost compared with almost previous methods, and it achieves less speed cost with increasing image resolution compared to the latest Vit-based image architecture Uformer.}
    \label{speed and gflops}
% \\
% \centering
% \includegraphics[width=45mm]{figure/sots_psnr.png} \label{fig:flops}

\end{figure}

Inexpensive computation is introduced above, we further explore the merits of the overall architecture for Dual-former. We present the GFLOPs and time cost with respect to different image sizes in Fig.\ref{speed and gflops}. The growth rates of them are not sensitive to image resolutions compared with other SOTA methods such as Uformer~\cite{wang2022uformer} and MPRNet~\cite{mpr}. In other words, embedding ViT-based modules into hidden layers improves model performance without significantly increasing computational and time cost as the resolution changes.
We can also observe that our speed and computational complexity are almost the lowest compared to the SOTA methods, demonstrating that our overall architecture achieves a balance between performance and computation for image restoration tasks and is more efficient for processing high-resolution images.
% \begin{figure}[!h]
%     \centering
%     \includegraphics[width=8cm]{figure/gflops.png}
%     \caption{GFLOPs of our method and SOTA methods with respect to different image resolutions. Our Dual-former has fewer GFLOPs than all competitors, and it enjoys the lowest GFLOPs growth rate inspired by input image size.}
%     \label{gflops}
% \end{figure}
% \vspace{0cm}
% \begin{figure}[!h]
%     \centering
%     \includegraphics[width=8cm]{figure/speed.png}
%     \caption{Time cost corresponding to various input image sizes of proposed Dual-former and SOTA approaches. Dual-former is superior in time cost compared with almost previous methods, and it achieves less speed cost with increasing image resolution compared to the latest Vit-based image architecture Uformer.}
%     \label{speed}
% \end{figure}

\subsection{Quantifiable Performance for High-level Vision Tasks}
As shown in Fig.\ref{highlevel}, we present the detection confidences supported by Google Vision API\footnote{Google Vision API : \url{https://cloud.google.com/vision/}}. From Fig.\ref{highlevel}, we notice that the degraded information affects the detection results to some extent. Specifically, haze, rain and snow reduce the objection numbers and confidences of detection, which limits the performance of high-level vision tasks. After processing by Dual-former, the restored image can tackle this problem and promote the development of high-level tasks.

%------------------------------------Ablation Study
\subsection{Ablation Study}
In this section, we conduct extensive ablation studies to verify the effect of our Dual-former network design. Specifically, we adopt the 192$\times$192 patch size of images on CSD~\cite{hdcwnet}  training set as input and train 200 epochs. Then we observe the results on CSD~\cite{hdcwnet} testing dataset to analyze crucial components.
We illustrate the influence of each composition in the following parts individually.

% \begin{figure*}[!t]
% \centering
% \subfloat[\scriptsize PSNR comparison on SOTS-Indoor dataset]{
% \includegraphics[width=57mm]{figure/SOTS-Indoorflops_psnr.png} \label{hazeflops}}
% \hfill
% \subfloat[\scriptsize PSNR comparison on Rain100L dataset]{
% \includegraphics[width=57mm]{figure/Rain100Lflops_psnr.png} \label{rainflops}}
% \hfill
% \subfloat[\scriptsize PSNR comparison on CSD dataset]{
% \includegraphics[width=57mm]{figure/CSDgflops_psnr.png} \label{snowflops}}
% \caption{Trade-off between performance vs computation complexity (GFLOPs) on three datasets of single image dehazing, deraining and desnowing. The results show that our Dual-former outperforms other state-of-the-art methods within small computation overhead.}\label{gflop}

% \end{figure*}

%--------------------------------------desnow visual comparison

%-------------------------------------Effectiveness of Channel-Spatial Self-attention

\subsubsection{Effectiveness of Channel-Spatial Self-attention} We first verify the effect of our Channel-Spatial Self-attention, which differs from other single dimensional perspective self-attention such as channel level~\cite{zamir2021restormer} or spatial level~\cite{vit}. We perform the following experiments on validating the superior of multi-perspective self-attention compared with other self-attention. (a) We replace the Channel-Spatial Self-attention with the self-attention module inspired by vanilla Transformer~\cite{vit} \textbf{(SSA)}. (b) we explore the gain of depth-wise convolution in spatial self-attention operation.(c) We apply self-attention across channel~\cite{zamir2021restormer} to substitute our Hybrid Self-attention design \textbf{(CSA)}. (d) We use Channel-Spatial Self-attention operation in our architecture \textbf{(CSSA)}. Table \ref{sa ab} reports comparison results for different configurations. We observe that Channel-Spatial Self-attention can provide a better representation to recover degraded images, which we believe can be explained by global modeling from both channel and spatial levels are more powerful than single self-attention for image restoration. And the depth-wise can enhance the local bias in spatial self-information.

\begin{table}[!h]
\centering
\caption{Ablation Study on Self-attention of different dimensional perspectives. Underline indicates the best metrics. (PSNR(dB)/SSIM) }\label{sa ab}
\resizebox{8cm}{!}{
\renewcommand\arraystretch{1.1}
\begin{tabular}{cccccccc}
\toprule[1.2pt]
\textbf{Setting} & \textbf{Model} & \multicolumn{2}{c}{\textbf{\#Param}} & \multicolumn{2}{c}{\textbf{\#GFLOPs}} & \textbf{PSNR} & \textbf{SSIM}  \\
\midrule[0.15em]
a& SSA & \multicolumn{2}{c}{14.21M} & \multicolumn{2}{c}{9.27G} & 30.21 & 0.936\\
b& SSA wo DWConv & \multicolumn{2}{c}{14.21M} & \multicolumn{2}{c}{9.26G} & 30.06 & 0.933\\
c& CSA &  \multicolumn{2}{c}{14.26M} & \multicolumn{2}{c}{9.28G} & 30.36 & 0.937 \\
d & CSSA(Ours)&  \multicolumn{2}{c}{14.23M} & \multicolumn{2}{c}{9.28G} & \underline{30.64} & \underline{0.939}\\
\bottomrule[1.2pt]
\end{tabular}}
\end{table}
%-------------------------------------%Novelty for Multi-branch feed-forward network
\subsubsection{Novelty for Multi-branch feed-forward network (MBFFN)} In this article, we propose the MBFFN instead of Multilayer Perceptron (MLP)~\cite{swim} for image restoration of various severe weather datasets. To demonstrate that MBFFN outperforms previous feed-forward network (FFN) designs, we modify the multiple experiments of FFN, which can further prove the obtained gain from MBFFN. (a) We follow the MLP as FFN backbone just like~\cite{swim} \textbf{(MLP)}. (b) In the PVT~\cite{PVT}, authors present the ConvFFN to improve the local excavation capacity. We compare it in image restoration to observe their performance~\cite{lee2022knn} \textbf{(ConvFFN)}. (c) We also compare with LeFF module~\cite{wang2022uformer} in this part, which attracts the remarkable results in image restoration field \textbf{(LeFF)}. (d) We use MBFFN to replace the above designs \textbf{(MFFN)}. Table \ref{ffn ab} verifies that MBFFN can achieve non-trivial than other variants of FFN in transformer block. Our MBFFN cancels the FC layers and changes to apply convolution. Multi-branch operation enhances the multi-scale local representation ability, which is complementary to self-attention.

\begin{table}[!h]
\centering
\captionsetup{justification=raggedright}
\caption{Ablation Study on various configurations of Feed-forward Network. Underline indicates the best metrics. (PSNR(dB)/SSIM) }\label{ffn ab}
\resizebox{8cm}{!}{
\renewcommand\arraystretch{1.1}
\begin{tabular}{cccccccc}
\toprule[1.2pt]
\textbf{Setting} & \textbf{Model} & \multicolumn{2}{c}{\textbf{\#Param}} & \multicolumn{2}{c}{\textbf{\#GFLOPs}} & \textbf{PSNR} & \textbf{SSIM}  \\
\midrule[0.15em]
a & MLP~\cite{swim} & \multicolumn{2}{c}{14.56M} & \multicolumn{2}{c}{9.36G} & 30.37 & 0.937\\
b& ConvFFN~\cite{lee2022knn} &  \multicolumn{2}{c}{14.61M} & \multicolumn{2}{c}{9.37G} & 30.42 & 0.938 \\
c & LeFF~\cite{wang2022uformer}&  \multicolumn{2}{c}{14.61M} & \multicolumn{2}{c}{9.39G} & {30.53} & {0.938}\\
d & MFFN(Ours) &  \multicolumn{2}{c}{14.23M} & \multicolumn{2}{c}{9.28G} & \underline{30.64} & \underline{0.939}\\
\bottomrule[1.2pt]
\end{tabular}}
\end{table}

%-------------------------------------%Benefits of Adaptive Control Module
\subsubsection{Benefits of Adaptive Control Module} To enhance the interaction of hybrid self-attention (Channel-Spatial Self-attention), we carefully introduce the Adaptive Control Module (ACM), which offsets the shortcoming of different dimensional perspectives to fuse the information ideally. We conduct ablation studies to explore various ways of modulating and their gains. (a) Without any Adaptive Control Module, only using concatenate operation to substitute it \textbf{(Concat)}. (b) We follow the ~\cite{song2022vision} inspired by~\cite{li2019selective}, testing the SK fusion to fuse multiple branches.
(c) We use the Adaptive Control Module (ACM) to replace other modules.
Compared with SK fusion and concatenate operation, our ACM requires more computation and parameters, while instead, it has better performance in PSNR and SSIM. This increase is within the acceptable range compared to the computational cost. It demonstrates that our ACM improves the capability of fusing the information from the channel and spatial levels, leading to superior performance over SK fusion and simple concatenation.
\begin{table}[!t]
\centering
\captionsetup{justification=raggedright}
\caption{Ablation Study on different information fusion operations. Underline indicates the best metrics. (PSNR(dB)/SSIM) }\label{acm ab}
\resizebox{8cm}{!}{
\renewcommand\arraystretch{1.1}
\begin{tabular}{cccccccc}
\toprule[0.15em]
\textbf{Setting} & \textbf{Model} & \multicolumn{2}{c}{\textbf{\#Param}} & \multicolumn{2}{c}{\textbf{\#GFLOPs}} & \textbf{PSNR} & \textbf{SSIM}  \\
\midrule[0.15em]
a & Concat & \multicolumn{2}{c}{9.78M} & \multicolumn{2}{c}{8.76G} & 30.03 & 0.932\\
b & SK fusion~\cite{song2022vision} & \multicolumn{2}{c}{9.82M} & \multicolumn{2}{c}{8.76G} & 30.12 & 0.933\\
c & ACM(Ours) &  \multicolumn{2}{c}{14.23M} & \multicolumn{2}{c}{9.28G} & \underline{30.64} & \underline{0.939}\\
\bottomrule[0.15em]
\end{tabular}}
\end{table}
%-------------------------------------%Necessity of Channel Attention in Local Feature Extract Block
\subsubsection{Necessity of Channel Attention (CA) in Local Feature Extraction Block (LFE)} 
For the proposed LFE, we embed channel attention to consolidate the information exchange between channel dimensions, improving the overall performance in the encoder-decoder part. The detailed experiments are shown in Table \ref{ca ab}. It's worth that we notice that the channel attention brings a certain degree of effect, while The amount of computation and parameters only rise a little ($\mathbf{9.24G\rightarrow9.28G}$ and $\mathbf{12.52M\rightarrow14.28M}$). This conclusion reflects that channel attention plays a vital role in our purpose. 

\begin{table}[!h]
\centering
\captionsetup{justification=raggedright}
\caption{Ablation Study on Channel Attention in Local Feature Extraction module and Hybrid Transformer Block with LFE module. Underline indicates the best metrics. (PSNR(dB)/SSIM) } \label{ca ab}
\vspace{-2mm}
\setlength{\tabcolsep}{2pt}
\scalebox{1}{
\begin{tabular}{cccccc}
\toprule[0.15em]
\textbf{LFE w CA} & \textbf{$\#$Param} & \textbf{$\#$GFLOPs} & \textbf{PSNR} & \textbf{SSIM} \\
\XSolidBrush &12.52M & 9.24G & 30.22 & 0.934\\
\checkmark&14.23M & 9.28G & \underline{30.64} & \underline{0.939} \\
\toprule[0.15em]
\textbf{HTB w LFE} &\textbf{$\#$Param} & \textbf{$\#$GFLOPs} & \textbf{PSNR} & \textbf{SSIM} \\
\XSolidBrush &11.01M & 8.71G & 30.04 & 0.932\\
\checkmark &14.23M & 9.28G & \underline{30.64} & \underline{0.939} \\
\bottomrule[0.15em]
\end{tabular}}
\end{table}
%--------------------------------------Benefits of Local Feature Extraction
\subsubsection{Benefits of Local Feature Extraction in Hybrid Transformer Block}
In the latent layer of the network, we design a parallel architecture to utilize a subset of channels for our local feature extraction. It can complement another part of the global modeling information to enhance the ability in the deepest layer, thereby improving the model's overall performance. In this ablation experiment, we demonstrate the effectiveness of this design. It can be seen from Table \ref{ca ab} that this parallelized design dramatically enhances the overall image restoration capability of the model and maintains a small computational cost.

% \begin{table}[!h]
% \centering
% %\setlength{\abovecaptionskip}{0.7cm}
% \captionsetup{justification=raggedright}
% \caption{Ablation Study on Hybrid Transformer Block with LFE module. Underline indicates the best metrics. (PSNR(dB)/SSIM) } \label{htb lfe}
% \resizebox{8cm}{!}{
% \renewcommand\arraystretch{1.1}
% \begin{tabular}{cccccc}
% \toprule[0.15em]
% \textbf{HTB w LFE} &\textbf{$\#$Param} & \textbf{$\#$GFLOPs} & \textbf{PSNR} & \textbf{SSIM} \\
% \midrule[0.15 em]
% \XSolidBrush &11.01M & 8.71G & 30.04 & 0.932\\
% \checkmark &14.23M & 9.28G & \underline{30.64} & \underline{0.939} \\
% \bottomrule[0.1em]
% \end{tabular}}
% \end{table}
%---------------------------------Effectiveness of Loss Function
\subsubsection{Effectiveness of Loss Function} 
In this ablation study, we explore the effectiveness of our chosen loss function. We apply the PSNRloss~\cite{chen2021hinet} as our reconstruction loss instead of L1 loss. At the same time, we study to prove the effect of perceptual loss in our train setting. Table \ref{loss ab} describes that PSNRloss has a slight advantage over L1 loss, perceptual loss is helpful during training due to its supervision at the feature level.
\begin{table}[!h]
\centering
\captionsetup{justification=raggedright}
\caption{Ablation Study on the effect of the loss function. Underline indicates the best metrics. (PSNR(dB)/SSIM) }\label{loss ab}
\resizebox{7cm}{!}{
\renewcommand\arraystretch{1.1}
\begin{tabular}{cccc}
\toprule[0.15em]
\textbf{Setting} & \textbf{Model} & \textbf{PSNR} & \textbf{SSIM}  \\
\midrule[0.15em]
1& $L1+L_{p}$ & 30.57 & 0.939\\
2& $L_{psnr}$ & 30.34 & 0.936\\
3& $L_{p}+L_{psnr}$ (Ours) & \underline{30.64} & \underline{0.939}\\
\bottomrule[0.15em]
\end{tabular}}
\end{table}

\vspace{-0.7cm}
\section{Conclusion}
In this paper, we have proposed a powerful yet efficient image restoration backbone. In contrast to Transformer-based image restoration structures, we only embed the Hybrid Transformer Block into the latent layer to unleash its mighty global modeling capabilities. This design cleverly combines the ability of local extraction and global information interaction while possessing low computational complexity compared to existing large networks. Experiments show the superiority of our Dual-former consisting of Local Feature Extraction and Hybrid Transformer Block (Channel-Spatial Self-attention), which presents strong representation capability for different adverse weather restoration tasks. We hope our backbone can be used for other efficient and robust network designs and achieve impressive results in numerous fields, such as image super-resolution and image demoiring. 
\bibliographystyle{IEEEtranS}
\bibliography{sampleBibFile}

\vfill

\end{document}